%% file: MAIN.tex
\documentclass{article}

\usepackage[preprint]{neurips_2025}

\usepackage{booktabs}
\usepackage{longtable}

\usepackage[utf8]{inputenc} %
\usepackage[T1]{fontenc}    %
\usepackage{hyperref}       %
\usepackage{url}            %
\usepackage{booktabs}       %
\usepackage{amsfonts}       %
\usepackage{nicefrac}       %
\usepackage{microtype}      %
\usepackage[table]{xcolor}

\usepackage[pdftex]{graphicx}
\usepackage{listliketab}
\usepackage{hyperref}
\hypersetup{
    colorlinks=true,
    linkcolor=blue,
    filecolor=magenta,
    urlcolor=blue,
    citecolor=blue
    }
\usepackage{amsmath}
\usepackage{ulem} %
\usepackage{cleveref} %
\usepackage{todonotes}
\usepackage[section]{placeins}

\usepackage[frozencache,cachedir=.]{minted} 

\setminted{
  frame=none,
  bgcolor=gray!20,
  fontsize=\footnotesize,
  linenos,
  breaklines=true
} %

\usepackage{subcaption} %

\usepackage{float}
\floatstyle{plaintop}
\restylefloat{table}
\usepackage[tableposition=top]{caption}

\title{LibriBrain: Over 50 Hours of Within-Subject MEG to Improve Speech Decoding Methods at Scale}

\author{%
Miran \"{O}zdogan$^{1}$ 
\quad Gilad Landau$^{1}$ 
\quad Gereon Elvers$^{1}$ 
\quad Dulhan Jayalath$^{1}$ 
\AND Pratik Somaiya$^{1}$ 
\quad Francesco Mantegna$^{1,2}$ 
\quad Mark Woolrich$^{2}$ 
\quad Oiwi Parker Jones$^{1,2}$\\ \\
$^1$PNPL\includegraphics[height=1.5\fontcharht\font`\B]{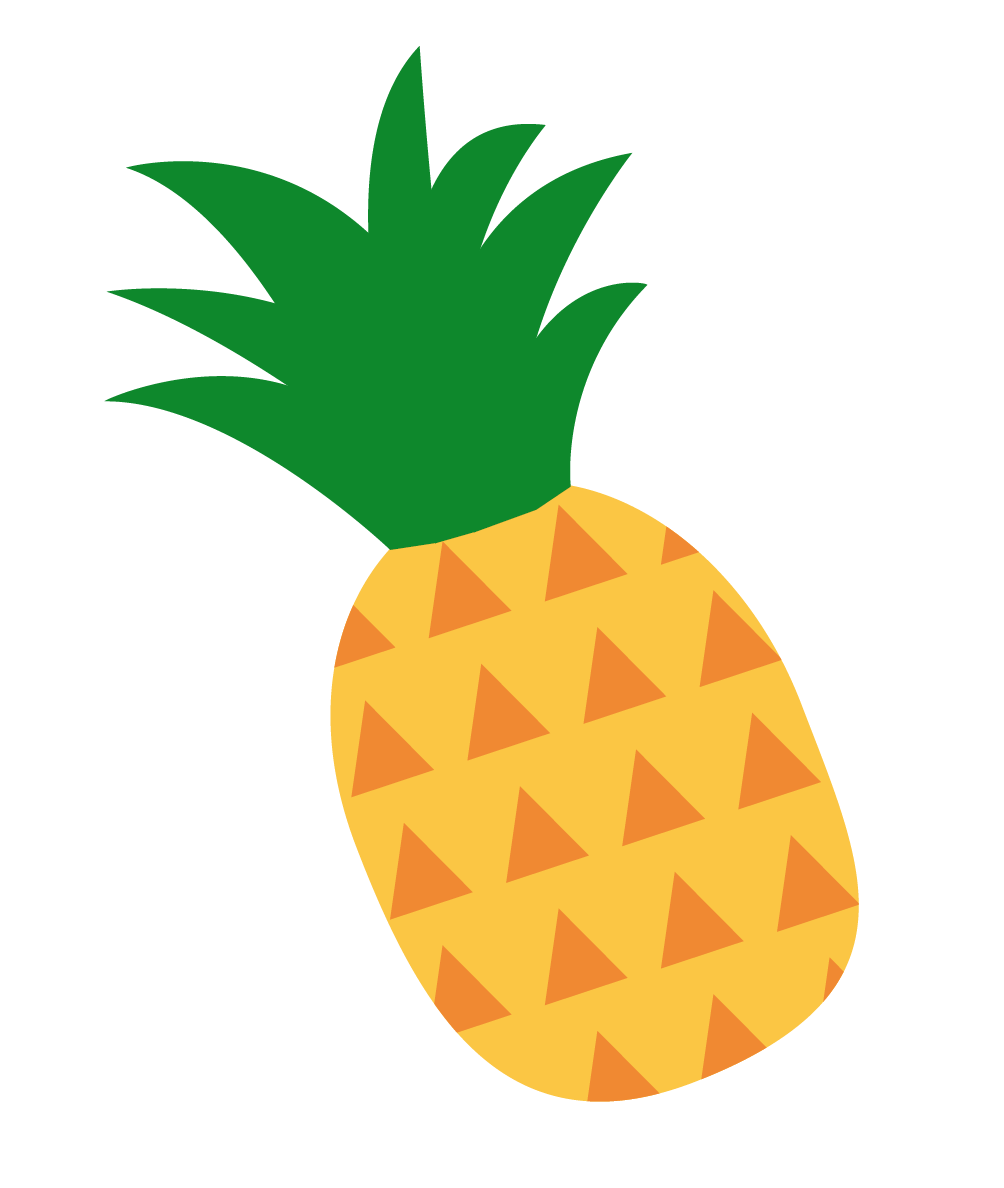}, Department of Engineering Science, University of Oxford, UK \\ \quad 
$^2$Oxford Centre for Human Brain Activity (OHBA), University of Oxford, UK \\
\\ \texttt{\{miran, oiwi\}@robots.ox.ac.uk}
}

\begin{document}

\maketitle
\begin{abstract}
\input{0_abstract}
\end{abstract}

\input{1_intro}
\input{2_related_work}

\input{tables/datasets}

\input{3_dataset}

\input{5_experiments}

\input{figures/scaling_all}
\input{5.1_speech_detection}

\input{tables/speech_detection_performance}

\input{5.2_phoneme_classification}

\input{tables/phoneme_classification_performance}

\input{5.3_word_classification}

\input{tables/word_classification_performance}

\input{6_discussion}

\input{6.0_broader_impacts}
\input{6.1_limitations}

\input{7_conclusion}

\input{8_ack}

\clearpage

\bibliography{references}
\bibliographystyle{apalike}

\newpage

\input{999_appendices}

\section{Further Discussion}\label{appendix:further-discussion}
\input{6.2_ethical_considerations}

\input{4_methods}

\input{4.2_stimuli}

\input{4.3_task}

\input{4.4_protocol}

\input{4.5_preprocessing}

\input{4.6_events}

\input{4.7_splits}

\input{supplements/extended_dataset_stuff}
\input{supplements/phoneme_classification}
\input{supplements/speech_detection}
\input{supplements/word_classification}

\newpage
\section*{NeurIPS Paper Checklist}

The checklist is designed to encourage best practices for responsible machine learning research, addressing issues of reproducibility, transparency, research ethics, and societal impact. Do not remove the checklist: {\bf The papers not including the checklist will be desk rejected.} The checklist should follow the references and follow the (optional) supplemental material.  The checklist does NOT count towards the page
limit. 

Please read the checklist guidelines carefully for information on how to answer these questions. For each question in the checklist:
\begin{itemize}
    \item You should answer \answerYes{}, \answerNo{}, or \answerNA{}.
    \item \answerNA{} means either that the question is Not Applicable for that particular paper or the relevant information is Not Available.
    \item Please provide a short (1–2 sentence) justification right after your answer (even for NA). 
\end{itemize}

{\bf The checklist answers are an integral part of your paper submission.} They are visible to the reviewers, area chairs, senior area chairs, and ethics reviewers. You will be asked to also include it (after eventual revisions) with the final version of your paper, and its final version will be published with the paper.

The reviewers of your paper will be asked to use the checklist as one of the factors in their evaluation. While "\answerYes{}" is generally preferable to "\answerNo{}", it is perfectly acceptable to answer "\answerNo{}" provided a proper justification is given (e.g., "error bars are not reported because it would be too computationally expensive" or "we were unable to find the license for the dataset we used"). In general, answering "\answerNo{}" or "\answerNA{}" is not grounds for rejection. While the questions are phrased in a binary way, we acknowledge that the true answer is often more nuanced, so please just use your best judgment and write a justification to elaborate. All supporting evidence can appear either in the main paper or the supplemental material, provided in appendix. If you answer \answerYes{} to a question, in the justification please point to the section(s) where related material for the question can be found.

IMPORTANT, please:
\begin{itemize}
    \item {\bf Delete this instruction block, but keep the section heading ``NeurIPS Paper Checklist"},
    \item  {\bf Keep the checklist subsection headings, questions/answers and guidelines below.}
    \item {\bf Do not modify the questions and only use the provided macros for your answers}.
\end{itemize}

\begin{enumerate}

\item {\bf Claims}
    \item[] Question: Do the main claims made in the abstract and introduction accurately reflect the paper's contributions and scope?
    \item[] Answer: \answerYes{} %
    \item[] Justification: The claims made in the abstract and introduction accurately reflect the properties of the dataset as described in Section~\ref{sec:dataset}, the experimental findings detailed in Section~\ref{sec:experiments}, and the limitations discussed in Section~\ref{sec:limitations}.

\item {\bf Limitations}
    \item[] Question: Does the paper discuss the limitations of the work performed by the authors?
    \item[] Answer: \answerYes{} %
    \item[] Justification: We discuss limitations in Section~\ref{sec:limitations}. Ethical and privacy considerations are discussed in Appendix~\ref{appendix:ethics}.

\item {\bf Theory assumptions and proofs}
    \item[] Question: For each theoretical result, does the paper provide the full set of assumptions and a complete (and correct) proof?
    \item[] Answer: \answerNA{}
    \item[] Justification: We do not present theoretical results. 

    \item {\bf Experimental result reproducibility}
    \item[] Question: Does the paper fully disclose all the information needed to reproduce the main experimental results of the paper to the extent that it affects the main claims and/or conclusions of the paper (regardless of whether the code and data are provided or not)?
    \item[] Answer: \answerYes{}
    \item[] Justification: Aside from providing the code we also detail all information necessary to reproduce of our experimental results in Appendix~\ref{appendix:phoneme},~\ref{appendix:speech}, and,~\ref{appendix:word-class}. We also provide simple access to the dataset ensuring that our results are easily reproducible.

\item {\bf Open access to data and code}
    \item[] Question: Does the paper provide open access to the data and code, with sufficient instructions to faithfully reproduce the main experimental results, as described in supplemental material?
    \item[] Answer: \answerYes{}
    \item[] Justification: We provide execution-ready scripts to recreate the training runs of the experimental results. The link to these is available in Section~\ref{sec:experiments}.

\item {\bf Experimental setting/details}
    \item[] Question: Does the paper specify all the training and test details (e.g., data splits, hyperparameters, how they were chosen, type of optimizer, etc.) necessary to understand the results?
    \item[] Answer: \answerYes{}
    \item[] Justification: We detail information necessary to understand the main results in Section~\ref{sec:experiments}. We describe data splits in Section~\ref{sec:data-splits}. We provide the full training and test details in Appendix~\ref{appendix:phoneme},~\ref{appendix:speech}, and~\ref{appendix:word-class}.

\item {\bf Experiment statistical significance}
    \item[] Question: Does the paper report error bars suitably and correctly defined or other appropriate information about the statistical significance of the experiments?
    \item[] Answer: \answerYes{}
    \item[] Justification: We report error bars that represent standard error across multiple seeds in Section~\ref{sec:experiments}. For the phoneme classification and speech detection experiments we report statistical significance compared to a naive baseline as measured by an exact permutation test. 

\item {\bf Experiments compute resources}
    \item[] Question: For each experiment, does the paper provide sufficient information on the computer resources (type of compute workers, memory, time of execution) needed to reproduce the experiments?
    \item[] Answer: \answerYes{}
    \item[] Justification: We provide details on computer resources need to reproduce the results in Appendix~\ref{appendix:speech:compute},~\ref{appendix:phoneme_compute}, and~\ref{appendix:word_compute}.
    
\item {\bf Code of ethics}
    \item[] Question: Does the research conducted in the paper conform, in every respect, with the NeurIPS Code of Ethics \url{https://neurips.cc/public/EthicsGuidelines}?
    \item[] Answer: \answerYes{} %
    \item[] Justification: We discuss ethical considerations at length in \ref{appendix:further-discussion}. The NeurIPS Code of Ethics was reviewed and adhered to.

\item {\bf Broader impacts}
    \item[] Question: Does the paper discuss both potential positive societal impacts and negative societal impacts of the work performed?
    \item[] Answer: \answerYes{}
    \item[] Justification: We provide an extensive discussion of potential positive and negative societal impacts in Section~\ref{sec:discussion}, and Appendix~\ref{appendix:further-discussion}.
    
\item {\bf Safeguards}
    \item[] Question: Does the paper describe safeguards that have been put in place for responsible release of data or models that have a high risk for misuse (e.g., pretrained language models, image generators, or scraped datasets)?
    \item[] Answer: \answerNA{}
    \item[] Justification: There are no immediate risks for misuse associated with the data and models we are publishing. However, we are intentionally releasing them under a CC BY-NC license to retain the ability to prevent potential unethical commercial use in the future.

\item {\bf Licenses for existing assets}
    \item[] Question: Are the creators or original owners of assets (e.g., code, data, models), used in the paper, properly credited and are the license and terms of use explicitly mentioned and properly respected?
    \item[] Answer: \answerYes{}
    \item[] Justification: The baseline word classification results rely on the dataset published by \citet{armeni2022}. We have appropriately credited the authors in the main text and explicitly referenced the dataset's license (RU-DI-HD-1.0) in Appendix~\ref{appendix:word-class}.


\item {\bf New assets}
    \item[] Question: Are new assets introduced in the paper well documented and is the documentation provided alongside the assets?
    \item[] Answer: \answerYes{}
    \item[] Justification: We provide extensive documentation alongside our code and dataset on GitHub and Hugging Face including information on licensing, training, and limitations.

\item {\bf Crowdsourcing and research with human subjects}
    \item[] Question: For crowdsourcing experiments and research with human subjects, does the paper include the full text of instructions given to participants and screenshots, if applicable, as well as details about compensation (if any)? 
    \item[] Answer: \answerYes{}
    \item[] Justification: In Appendix~\ref{appendix:experimental_design} and~\ref{appendix:data_aquisition} we give a detailed account of instructions given to participants. 

\item {\bf Institutional review board (IRB) approvals or equivalent for research with human subjects}
    \item[] Question: Does the paper describe potential risks incurred by study participants, whether such risks were disclosed to the subjects, and whether Institutional Review Board (IRB) approvals (or an equivalent approval/review based on the requirements of your country or institution) were obtained?
    \item[] Answer: \answerYes{}
    \item[] Justification: The participant provided informed consent for data collection and approved the sharing of pseudonymised data for research purposes, in accordance with the University of Oxford's ethical oversight. 

\item {\bf Declaration of LLM usage}
    \item[] Question: Does the paper describe the usage of LLMs if it is an important, original, or non-standard component of the core methods in this research? Note that if the LLM is used only for writing, editing, or formatting purposes and does not impact the core methodology, scientific rigorousness, or originality of the research, declaration is not required.
    \item[] Answer: \answerNA{}
    \item[] Justification: This research does not involve LLMs as any important, original, or non-standard components.

\end{enumerate}
\end{document}

%% file: 0_abstract.tex
LibriBrain represents the largest single-subject MEG dataset to date for speech decoding, with over 50 hours of recordings---5$\times$ larger than the next comparable dataset and 50$\times$ larger than most. This unprecedented `depth' of within-subject data enables exploration of neural representations at a scale previously unavailable with non-invasive methods. LibriBrain comprises high-quality MEG recordings together with detailed annotations from a single participant listening to naturalistic spoken English, covering nearly the full Sherlock Holmes canon. Designed to support advances in neural decoding, LibriBrain comes with a Python library for streamlined integration with deep learning frameworks, standard data splits for reproducibility, and baseline results for three foundational decoding tasks: speech detection, phoneme classification, and word classification. Baseline experiments demonstrate that increasing training data yields substantial improvements in decoding performance, highlighting the value of scaling up deep, within-subject datasets. By releasing this dataset, we aim to empower the research community to advance speech decoding methodologies and accelerate the development of safe, effective clinical brain-computer interfaces.

%% file: 1_intro.tex
\input{figures/size}

\section{Introduction}\label{sec:1_intro}
Recent advances in deep learning have demonstrated the critical role of high-quality datasets in enabling robust and generalisable models — a principle that holds true not only in vision and language domains, but increasingly in the study of speech decoding from brain signals. In invasive brain-computer interface (BCI) research, large-scale datasets collected via electrocorticography (ECoG) or microelectrode arrays have made it possible to train models with dozens hours of data per subject, leading to remarkable performance in paralysed individuals \citep{moses2016, metzger2023neuroprosthesis, willett2023high}. Most notably, recent systems have achieved less than 5\% word error rates (WER) for text decoding in paralysed patients \citep{card2024nejm}.

However, these achievements come with a fundamental limitation: they rely on data obtained through brain surgery. While neural interfaces based on ECoG or microelectrode arrays are extremely powerful, the need for invasive recording presents a substantial barrier to deployment outside of clinical trials. This has placed renewed emphasis on the search for non-invasive alternatives, particularly those using EEG or MEG, to achieve high-quality speech decoding without requiring surgical intervention.

Although full non-invasive brain-to-text systems have yet to achieve WERs below 100\% \citep{jo2024eegtotext,yang2024neuspeech,yang2024mad,yang2024neugpt}, a growing body of work has shown strong results on intermediate tasks such as speech segment identification \citep{defossez2023}, word-level classification \citep{dascoli2024}, and feature decoding \citep{jayalath2024scaling, jayalath2025icml}. Crucially, these advances have been enabled by the increasing availability of open EEG and MEG datasets, which have lowered the barrier to training powerful models across diverse tasks and conditions.

Early studies focused on pooling data across subjects within a single dataset \citep{defossez2023,dascoli2024}, but recent work has begun to explore more ambitious scaling via cross-dataset pre-training \citep{jayalath2024scaling, jayalath2025icml} and domain adaptation \citep{ridge2024domainshift}. These findings establish a basis for future research aimed at leveraging extensive pooled non-invasive datasets to develop general-purpose decoding models. However, an important insight has recently emerged: although training on broader datasets with more subjects tends to improves generalisation, models trained on `deep' data from a single subject often outperform broader models when matched for total training hours\citep{dascoli2024}.

Most existing non-invasive datasets are `broad but shallow', offering 1–2 hours per subject. The standout exception is the dataset by \citet{armeni2022}, with over 10 hours per subject, which has underpinned some of the strongest decoding results to date \citep{dascoli2024}. This highlights an important gap in available resources, as non-invasive speech datasets with substantial per-subject depth remain limited yet seem crucial for progress toward reliable brain-to-text decoding.

\input{figures/teaser}

In this paper, we introduce LibriBrain, a new open-access magnetoencephalography (MEG) dataset recorded from a single healthy volunteer listening to naturalistic, connected speech.
With over 50 hours of data, LibriBrain offers the deepest non-invasive speech brain recordings to date, being more than 5× larger than the previous largest within-subject MEG dataset \citep{armeni2022}, and 50× larger than most datasets in this space (see Figure \ref{fig:size}). Inspired by the LibriSpeech dataset for automatic speech recognition (ASR), LibriBrain includes excerpts from seven audiobooks. All source audio is from LibriVox, ensuring public-domain availability and full reproducibility.

Speech decoding tasks can target overt, covert (inner), or heard speech. Of these, heard speech is the most tractable for large-scale non-invasive recording. It not only supports fundamental research in auditory representation and inner speech (e.g. as auditory imagery), but also aids in developing robust, transferable methods for speech BCIs. LibriBrain was designed to fill this methodological gap: enabling reproducible, scalable research on decoding speech from brain activity using naturalistic stimuli.

LibriBrain makes the following contributions:
\begin{itemize}
    \item \textbf{Largest within-subject MEG speech dataset to date}: Over 50 hours of high-quality MEG data from a single participant, exceeding previous datasets by 5–50×.
    \item \textbf{Naturalistic, richly annotated stimuli}: Recordings obtained while listening to full-length audiobooks, including six Sherlock Holmes books and a subset of LibriSpeech excerpts.
    \item \textbf{Designed for ML usability}: Easily accessible via \texttt{pip}, with a simple-to-use Python API (\texttt{from pnpl.datasets import LibriBrain}) and standard train/val/test splits.
    \item \textbf{Benchmark-ready}: Supports two initial benchmark tasks—speech detection and phoneme classification—and includes baseline models and code.
    \item \textbf{Community engagement}: Supports open ML competitions and a public leaderboard to encourage community-driven progress and reproducibility.
    \item \textbf{Open and reproducible}: All audio, transcripts, and metadata are shared under public-domain licenses, making the entire dataset freely usable and shareable.
\end{itemize}

%% file: figures/size.tex
\begin{figure}[htbp]
    \centering
\includegraphics[width=\linewidth]{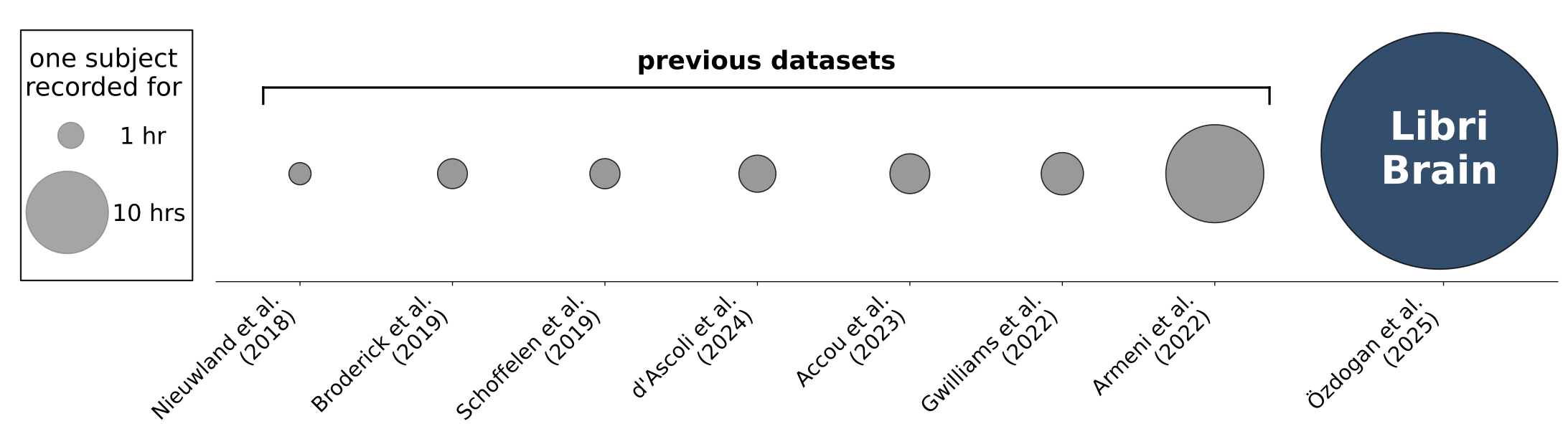}
    \caption{Comparison of within-subject data volume in non-invasive speech datasets.}
    \label{fig:size}
\end{figure}

%% file: figures/teaser.tex
\begin{figure}[htbp]
    \centering
\includegraphics[width=\linewidth]{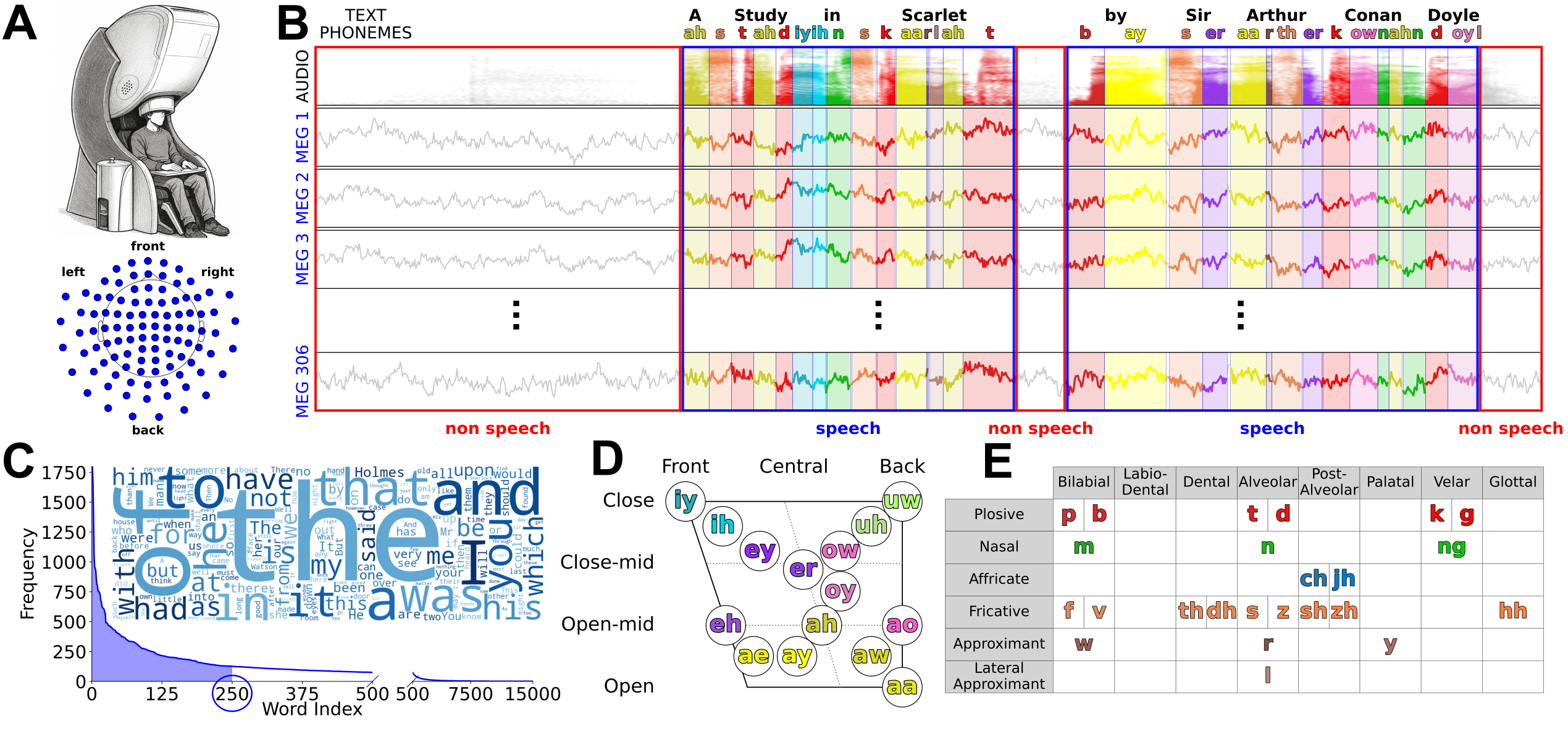}
    \caption{Overview of the LibriBrain dataset. (A) Illustration of the MEG scanner and sensor layout. (B) Example annotation of a sentence, showing phonemes and words aligned with the audio spectrogram and MEG time series. (C) Word frequency distribution and word cloud for the 250 most frequent words. (D) Vowel chart and (E) consonant chart depicting the linguistic properties of the phonemes.
    }
    \label{fig:teaser}
\end{figure}

%% file: 2_related_work.tex
\section{Related Work}\label{sec:2_related_work}

A growing number of EEG and MEG datasets focus on language processing and can be leveraged to scale non-invasive speech decoding models (Table \ref{tab:datasets}). These datasets vary widely across multiple dimensions, including sensor count, linguistic richness, recording duration, and the number of participants.

In terms of sensor coverage, the largest are LibriBrain and the MEG datasets of \citet{dascoli2024}, all recorded using 306-channel Elekta/MEGIN systems—matching the highest-density sensor arrays available for non-invasive human neuroimaging. 

When measured by total duration, the largest dataset is \citet{nieuwland2018} with 171 hours of EEG. However, this dataset is very broad and shallow: its 334 participants each contributed on average only 30 minutes of data, limiting per-subject decoding depth. Similar trends appear in the MEG datasets of \citet{schoffelen2019}, which span 81 hours (listening) and 106 hours (reading) but include over 90 subjects, with just 0.8–1.1 hours recorded per individual. In contrast, the dataset of \citet{armeni2022} is deeper, with approximately 10 hours per subject and 30–34 hours total, and has been shown to yield stronger decoding performance in follow-up studies \citep{dascoli2024}.

LibriBrain extends this direction by offering over 50 hours of MEG from a single subject—approximately 5× deeper than the next closest MEG dataset. While \citet{armeni2022} focused on one Sherlock Holmes book, LibriBrain spans nearly the entire Conan Doyle canon, with recordings from seven audiobooks. This depth allows for analysis of fine-grained speech representations at scale, and supports strong generalisation to unseen data in downstream decoding tasks (Section \ref{sec:experiments}).

Unlike prior work, LibriBrain is designed with machine learning usability in mind. It is fully open, publicly available on Hugging Face, and comes with a Python API (\texttt{pnpl}) for seamless integration with deep learning frameworks such as PyTorch \citep{paszke2019pytorch}. As shown in Table \ref{tab:datasets}, most recent datasets are public, but the raw MEG from \citet{dascoli2024} remains unavailable at the time of writing. Furthermore, LibriBrain includes ready-to-use benchmarks and reference models for speech detection, phoneme classification, and word decoding—supported by public leaderboards as part of the PNPL competition series—to enable standardised evaluation and accelerate progress in non-invasive speech BCI research.

Despite a growing number of non-invasive datasets and rising interest in foundation models for brain signals, current decoding models rarely generalise across datasets \citet{defossez2023, dascoli2024}. Historically, decoding has struggled to generalise across subjects, though individual subjects might benefit from group-level models \citet{csaky2023group}. However, a number of recent works have successfully pooled data across subjects, though often not between datasets. For example, although \citet{dascoli2024} analyse recordings from over 700 subjects, they report results separately for each dataset—underscoring a lack of cross-dataset generalisability. Recent work by \citet{jayalath2024scaling, jayalath2025icml} and \citet{ridge2024domainshift} shows that unsupervised pretraining and domain adaptation can help, but high-quality, within-subject data remains essential. LibriBrain takes a complementary approach by scaling data longitudinally within a single participant. It provides the deepest single-subject MEG dataset to date and is designed to reduce friction in neural decoding research, with standardised splits, public loaders, and baseline models to support reproducibility and benchmark-driven progress.

%% file: tables/datasets.tex
\begin{table}[tb]
\centering
\resizebox{\textwidth}{!}{%
\begin{tabular}{llllrrrrlll}
\toprule
\textbf{Dataset} & \textbf{Task} & \textbf{Modality} & \textbf{Language} & \textbf{Sensors} & \textbf{Total hrs} & \textbf{\# Subj} & \textbf{Hrs/Subj} & \textbf{Stimulus} & \textbf{Public} & \textbf{PNPL Dataloader} \\
\midrule
{LibriBrain} (ours) & \cellcolor{green!20}Listen & \cellcolor{green!20}MEG & \cellcolor{red!10}English & \textbf{306} & 52 & 1 & \textbf{52.32} & LibriVox (Sherlock, Books 1--7) & \cellcolor{green!20}\href{https://huggingface.co/datasets/pnpl/LibriBrain}{Yes} & \cellcolor{green!20}Yes \\
\citet{armeni2022} & \cellcolor{green!20}Listen & \cellcolor{green!20}MEG & \cellcolor{red!10}English & 269 & 30 & 3 & 10.0 & LibriVox (Sherlock, Book 3) & \cellcolor{green!20}\href{https://data.ru.nl/collections/di/dccn/DSC_3011085.05_995}{Yes} & \cellcolor{green!20}Yes (staged for release) \\
\citet{gwilliams2022natcomm} & \cellcolor{green!20}Listen & \cellcolor{green!20}MEG & \cellcolor{red!10}English & 208 & 49 & 27 & 1.0 & MASC stories, synthetic voice & \cellcolor{green!20}\href{https://doi.org/10.17605/OSF.IO/AG3KJ}{Yes} & \cellcolor{green!20}Yes (staged for release)\\
\citet{schoffelen2019} & \cellcolor{green!20}Listen & \cellcolor{green!20}MEG & \cellcolor{red!10}English & 275 & 81 & 96 & 0.8 & Spoken sentences & \cellcolor{green!20}\href{https://data.ru.nl/collections/di/dccn/DSC_3011020.09_236}{Yes} & \cellcolor{green!20}Yes (staged for release)\\
\citet{dascoli2024} & \cellcolor{green!20}Listen & \cellcolor{green!20}MEG & \cellcolor{blue!10}French & \textbf{306} & 94 & 58 & 1.6 & Le Petit Prince & \cellcolor{gray!20}No & \cellcolor{gray!20}Not yet \\
\citet{brennan2023alice} & \cellcolor{green!20}Listen & \cellcolor{gray!20}EEG & \cellcolor{red!10}English & 61 & 10.1 & 49 & 0.2 & Alice in Wonderland & \cellcolor{green!20}\href{https://deepblue.lib.umich.edu/data/concern/data_sets/bn999738r?locale=en}{Yes} & \cellcolor{gray!20}Not yet \\
\citet{dascoli2024} & \cellcolor{gray!20}Read & \cellcolor{green!20}MEG & \cellcolor{blue!10}French & \textbf{306} & 59 & 46 & 1.3 & Le Petit Prince & \cellcolor{gray!20}No & \cellcolor{gray!20}Not yet \\
\citet{schoffelen2019} & \cellcolor{gray!20}Read & \cellcolor{green!20}MEG & \cellcolor{orange!20}Dutch & 275 & 106 & 99 & 1.1 & Narrative reading task & \cellcolor{green!20}\href{https://data.ru.nl/collections/di/dccn/DSC_3011020.09_236}{Yes} & \cellcolor{gray!20}Not yet \\
\citet{accou2023} & \cellcolor{gray!20}Read & \cellcolor{gray!20}EEG & \cellcolor{red!10}English & 64 & 150 & 85 & 1.8 & Harry Potter & \cellcolor{green!20}\href{https://tudatalib.ulb.tu-darmstadt.de/handle/tudatalib/3914}{Yes} & \cellcolor{gray!20}Not yet \\
\citet{nieuwland2018} & \cellcolor{gray!20}Read & \cellcolor{gray!20}EEG & \cellcolor{red!10}English & 22 & \textbf{171} & \textbf{334} & 0.5 & Hemingway short story & \cellcolor{green!20}\href{https://osf.io/eyzaq/}{Yes} & \cellcolor{gray!20}Not yet \\
\citet{broderick2019} & \cellcolor{gray!20}Read & \cellcolor{gray!20}EEG & \cellcolor{red!10}English & 128 & 20 & 19 & 1.1 & Subset of \citet{nieuwland2018} & \cellcolor{green!20}\href{https://doi.org/10.5061/dryad.070jc}{Yes} & \cellcolor{gray!20}Not yet \\
\bottomrule
\\
\end{tabular}%
}
\caption{M/EEG datasets for language tasks.}
\label{tab:datasets}
\end{table}

%% file: 3_dataset.tex
\section{The LibriBrain Dataset}\label{sec:dataset}

\subsection{Dataset Overview}

The LibriBrain dataset consists of non-invasive magnetoencephalography (MEG) brain recordings, obtained from a single healthy volunteer while listening to over 50 hours of audiobook recordings. 
Paired event files contain time-locked annotations for linguistic events (e.g.~speech, words, and phonemes). 
The final dimensions of the data are 306 sensor channels $\times$ $T$ time samples, where each sample represents 4 milliseconds. At a high level of abstraction, the dataset is split into standard training, validation, and test sets to facilitate reproducible machine learning applications, with additional hidden sets reserved for competition evaluation \citep[see e.g.~][]{landau2025competition}.
At a lower level, the data are split into the experimental sessions in which they were collected, with whole sessions set aside as holdout data. 
Non-invasive MEG was collected at the Oxford Centre for Human Brain Activity (OHBA) using a state-of-the-art MEGIN Triux™ Neo system featuring 306 sensor channels that simultaneously probe magnetic fields across the entire head. The recordings were originally sampled at 1 kHz but were downsampled during preprocessing to 250 Hz to preserve oscillations into the high-gamma range (70--125 Hz).

\subsection{Data Collection and Structure}

MEG data were acquired for audiobook recordings of seven books in the canon of Sherlock Holmes (Table \ref{tab:train-audio} in Supplementary Materials). Recordings were made over 95 sessions, with each session corresponding to a book chapter (or occasionally part of a chapter when chapters were split into multiple parts in the audiobook recordings). Sessions vary in duration, with the average session lasting approximately 34 minutes, with a standard deviation of 15 minutes, corresponding to audiobook recordings with an average of 5421.28 words (standard deviation 2485.19).
Each recording session is paired with an event file (in CSV format) containing temporal information about linguistic events such as speech/non-speech, words, and phonemes; this information can be used for various encoding and decoding tasks, such as speech detection, phoneme classification, and word classification (see Section~\ref{sec:experiments}). 
Full experimental methods for data collection are in section \ref{sec:methods} of the Appendix. 

\input{tables/splits}

\subsection{Data Format, Access, and Supporting Python Library}

To lower the entry barrier for machine learning practitioners, we are initially releasing the data in serialised format (HDF5 files). Data are provided in a modular structure (one HDF5 file and TSV file per session), making it easy to load different amounts of data for training. We dedicate one session each for validation and testing (book 1 sessions 11 and 12, respectively). Two additional sessions are being held back for open machine learning competitions \citep{landau2025competition}. The validation, test, and competition holdout sessions were acquired on a separate day from all training data, to provide a strong protection against information leakage via `nonsense correlations' \citep{harris2021nonsense}.

For this release, we applied minimal preprocessing to maximise accessibility for ML practitioners unfamiliar with neuroimaging pipelines. Briefly, raw MEG recordings were corrected for head movement and signals were filtered (e.g.~to remove obvious noise) and then downsampled (see section \ref{sec:preproc} for details). Data can be accessed directly using standard HDF5 libraries. 
However, the recommended way to interact with the data is through our Python library \texttt{pnpl}:

\begin{minted}{python}
#!pip install pnpl  
from pnpl.datasets import LibriBrainSpeech, LibriBrainPhoneme
train_data = LibriBrainSpeech(path="./data", partition="train", download=True)
train_data = LibriBrainPhoneme(path="./data", partition="train")
\end{minted}

The initial release of the \texttt{pnpl} library supports two core tasks aligned with the 2025 PNPL competition: Speech Detection and Phoneme Classification. Nonetheless, the library design is modular, allowing for straightforward extensions to additional tasks derived from the accompanying events file. The source code of the \texttt{pnpl} library is available on GitHub.\footnote{\url{https://github.com/neural-processing-lab/frozen-pnpl}} The data is also available for direct download from Hugging Face.\footnote{\url{https://huggingface.co/datasets/pnpl/LibriBrain}}

For a future release, we are preparing a BIDS-compliant \citep{niso2018megbids} and pseudonymised version of the data in FIF format. This version of the data will take up significantly more memory than the serialised format, but include raw recordings for custom preprocessing. 
We plan to make it loadable through the \texttt{pnpl} library after the 2025 competition \citep{landau2025competition}.

%% file: tables/splits.tex
\begin{table}[tb]
\centering
\begin{tabular}{lrrrrrrr}
\toprule
\textbf{Split} & \textbf{Words} & \textbf{Unique Words} & \textbf{Phonemes} & \textbf{Sessions} & \textbf{Minutes} & \textbf{Hours} \\
\midrule
Train      & 459,227 & 16,753 & 1,488,392 & 91 & 3,094 & 51.57 \\
Validation & 3,427 & 1,082 & 11,289 & 1 & 22   & 0.36 \\
Test       & 3,576 & 1,145 & 12,051 & 1 & 23   & 0.38 \\
\midrule
\textbf{Total} & 466,230 & 16,892 & 1,511,732 & 93 & 3,139 & 52.32 \\
\bottomrule
\\
\end{tabular}
\caption{LibriBrain dataset statistics: word, phoneme, and audio duration counts across train, validation, and test splits. To reduce information leakage, the competition holdout data statistics are redacted here \citep[see][]{landau2025competition}.}
\label{tab:splits}
\end{table}

%% file: 5_experiments.tex
\section{Decoding Experiments}\label{sec:experiments}

To validate the LibriBrain dataset and establish strong baselines for future work, we evaluate three neural decoding tasks: speech detection, phoneme classification, and word classification. These span a spectrum of complexity, providing insights into the quality of the dataset and the scalability of neural speech decoding. All experiments are fully reproducible, using the standard train, validation, and test splits defined in the \texttt{pnpl} data loader. Code to reproduce these analyses is provided on GitHub.\footnote{\url{https://github.com/neural-processing-lab/libribrain-experiments}}

For speech detection and phoneme classification, we employ a lightweight convolutional neural network architecture inspired by SEANet \citep{seanet}, specifically adapted for MEG data input (Appendix~\ref{appendix:phoneme:hps}). To ensure meaningful baseline comparisons, we also evaluate performance against a random baseline that assigns labels according to their frequency in the training set.
For the word classification task, we replicate the state-of-the-art method introduced by \citet{jayalath2025cracking}.

%% file: figures/scaling_all.tex
\begin{figure}[tb]
  \centering
  \begin{subfigure}[b]{0.31\linewidth}
    \centering
    \includegraphics[width=\linewidth]{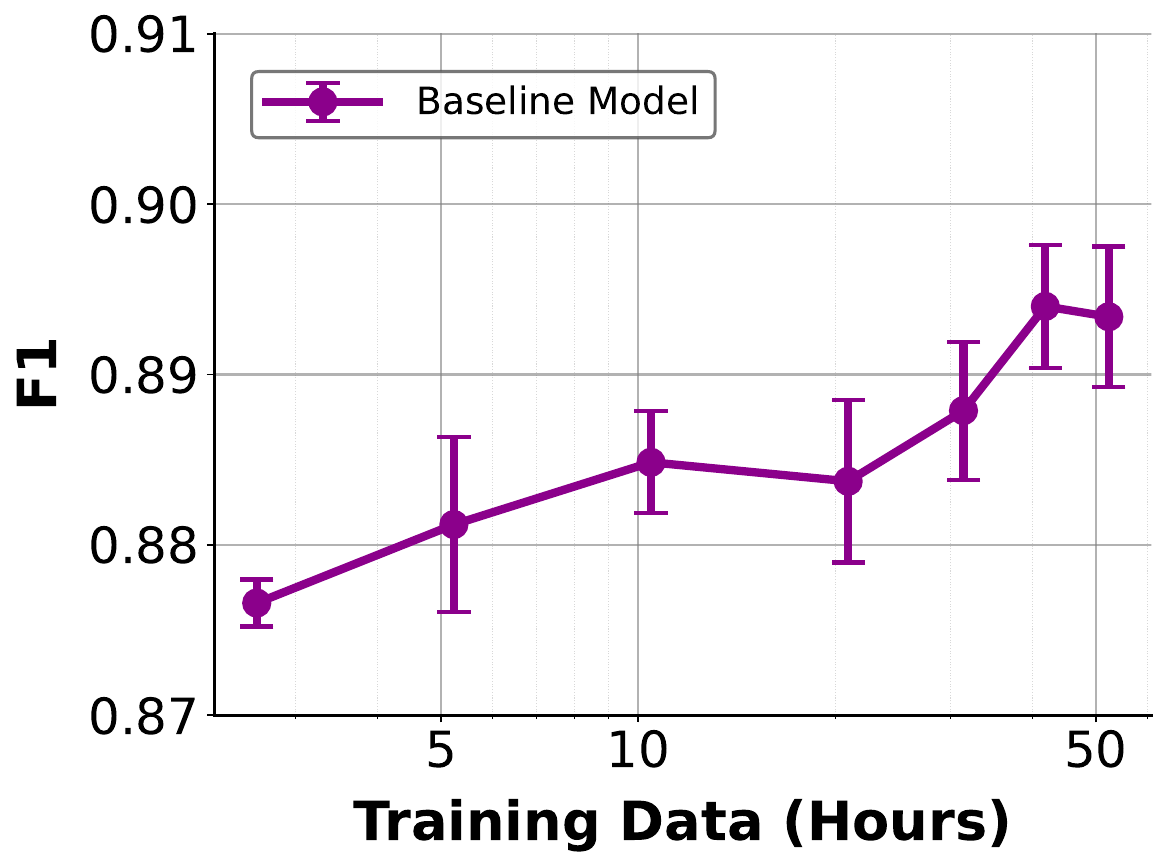}
    \caption{Speech detection}
    \label{fig:speech_scaling}
  \end{subfigure}
  \hfill
  \begin{subfigure}[b]{0.31\linewidth}
    \centering
    \includegraphics[width=\linewidth]{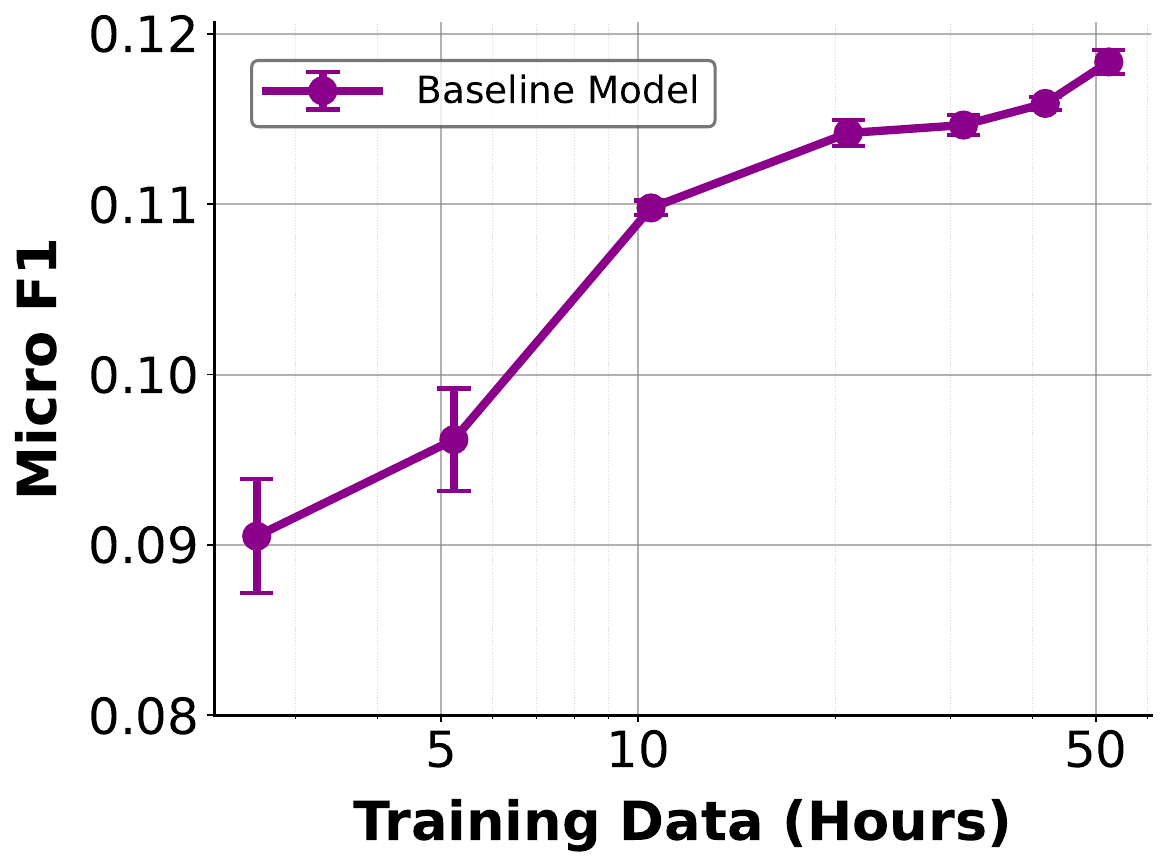}
    \caption{Phoneme classification}
    \label{fig:phoneme_scaling}
  \end{subfigure}
  \hfill
  \begin{subfigure}[b]{0.31\linewidth}
    \centering
    \includegraphics[width=\linewidth]{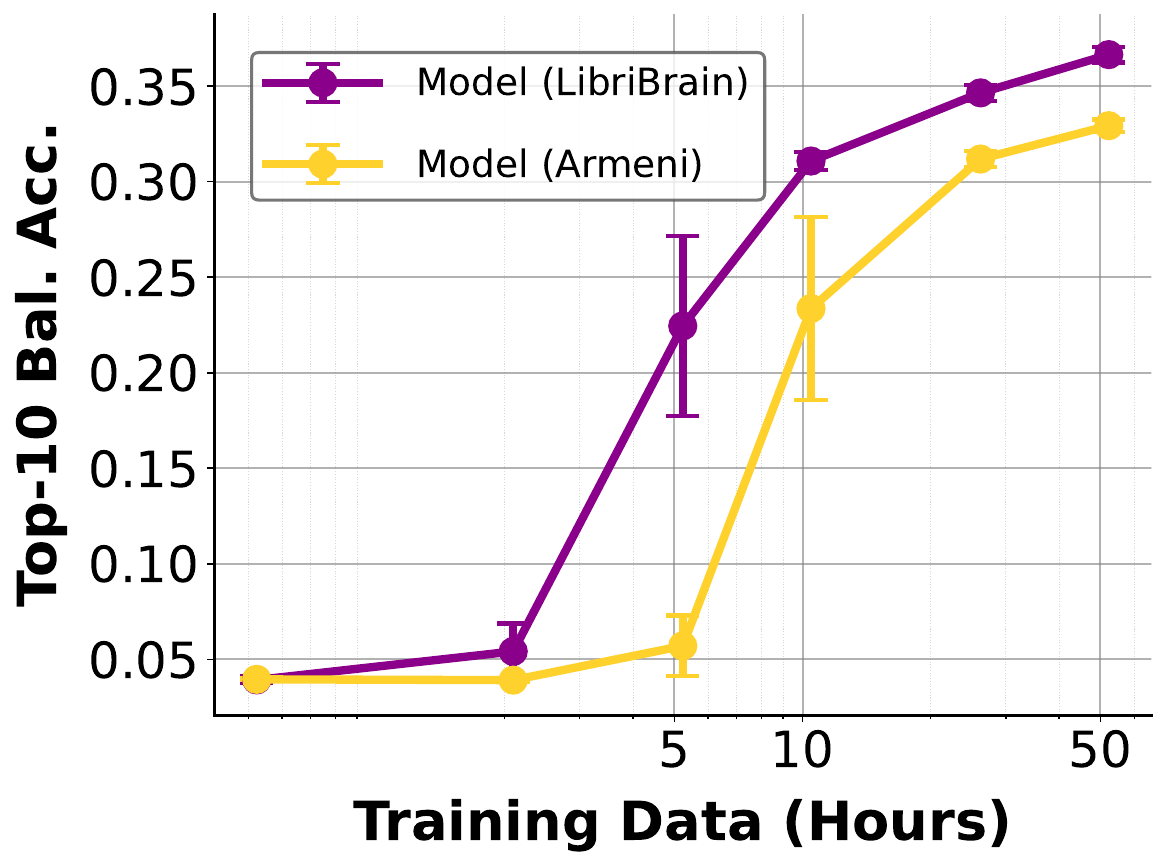}
    \caption{Word classification}
    \label{fig:word_scaling}
  \end{subfigure}
  \caption{Impact of training data volume on model performance across tasks. Subsets were randomly drawn from the training split. Results for speech detection and phoneme classification are averaged over three random seeds, with error bars showing standard error. Word classification results, reproduced with permission from \citet{jayalath2025cracking}, use five random seeds.}
  \label{fig:scaling_all}
\end{figure}

%% file: 5.1_speech_detection.tex
\subsection{Task 1: Speech Detection}

In the speech detection task, the objective is to accurately identify the time points within a segment of MEG recordings when participants were actively listening to speech.

We assess model performance using a number metrics: F1-Score, Balanced Accuracy, Area Under the Receiver Operating Characteristic curve (AUROC), Jaccard Index, and Cross Entropy Loss. We present the results in Table~\ref{tab:speech_detection_performance}.

Additionally, we analyse the relationship between the amount of data used for training and model performance, observing that performance improves approximately logarithmically with increasing amounts of training data (Figure~\ref{fig:speech_scaling}).

%% file: tables/speech_detection_performance.tex
\begin{table}[htb]
  \centering
  \begin{tabular}{lcc}
    \toprule
    \textbf{Metric} & \textbf{Model} & \textbf{ Random Baseline} \\
    \midrule
    F1-score &  \textbf{0.8989}$^*$ $\pm$ 0.0007 & 0.7848 $\pm$ 0.0017 \\
    Balanced Accuracy &  \textbf{0.7082}$^*$ $\pm$ 0.0031 & 0.4990 $\pm$ 0.0028 \\
    AUROC &  \textbf{0.8644}$^*$ $\pm$ 0.0017 & 0.5000 $\pm$ 0.0000 \\
    Jaccard Index &  \textbf{0.6077}$^*$ $\pm$ 0.0027 & 0.3820 $\pm$ 0.0024 \\
    Cross Entropy Loss &  \textbf{0.3802}$^*$ $\pm$ 0.0044 & 0.5449 $\pm$ 0.0000 \\
    \bottomrule
    \\
  \end{tabular}
  \caption{Speech detection performance. Mean and standard error of metrics over 10 seeds.} 
    {\footnotesize%
    $^{*}$ $p<0.01$, one‑sided exact permutation test using all 1,024 sign‑flip permutations.%
  }

\label{tab:speech_detection_performance}
  \end{table}

%% file: 5.2_phoneme_classification.tex
\subsection{Task 2: Phoneme Classification}

The goal of phoneme classification is to predict which of the 39 ARPAbet phonemes \citep{weide1998} corresponds to a given segment of MEG data. Each training sample consists of a short MEG window precisely aligned with a single phoneme from the stimulus audio. This task builds upon the extensive tradition of phoneme-based automatic speech recognition (ASR) \citep{garofolo1993timit}, and continues to play a central role in contemporary invasive brain-to-text pipelines \citep{wilson2020allphonemes, metzger2023neuroprosthesis, willett2023high, card2024nejm}.

We evaluate model performance using F1-Score, Balanced Accuracy, AUROC, Jaccard Index, and Cross Entropy Loss, with complete results presented in Table~\ref{tab:phoneme_hyperparameters}. For F1 score and AUROC, we report both micro- and macro-averaged values. The results demonstrate that the data enables statistically significant improvements in decoding performance over the random baseline. 
The absence of improvement in macro F1-score can be attributed to the phoneme frequency distribution, which roughly follows a power-law pattern; consequently, our model does not predict the less frequent phonemes, leading to penalties under this metric. In \citet{landau2025competition}, we employ a class-weighted loss function, resulting in statistically significant improvements in macro F1-score over the random baseline.
For a more comprehensive analysis, refer to Appendix~\ref{appendix:phoneme}.

Additionally, we investigate scaling behaviour, finding that decoding accuracy improves approximately logarithmically with additional training data (Figure \ref{fig:phoneme_scaling}), consistent with scaling laws in other ML domains \citep{llm_scaling_laws, image_scaling}.

Finally, we investigate the effects of averaging multiple instances of the same phoneme before classification. 
This procedure simulates brain-computer interface scenarios, where repeated user inputs are leveraged to enhance decoding reliability. 
We observe that accuracy increases consistently with the number of averaged tokens, achieving a balanced accuracy exceeding 60\% with 100 repetitions (Figure~\ref{fig:phoneme_averaging}).

%% file: tables/phoneme_classification_performance.tex
\begin{table}[htb]
  \centering
  \begin{tabular}{lcc}
    \toprule
    \textbf{Metric} & \textbf{Model} & \textbf{ Random Baseline} \\
    \midrule
    Micro F1-score &  \textbf{0.1168}$^*$ $\pm$ 0.0003 & 0.0442 $\pm$ 0.0003 \\
    Macro F1-score &  0.0253 $\pm$ 0.0005 & \textbf{0.0258} $\pm$ 0.0003 \\
    Balanced Accuracy &  \textbf{0.0399}$^*$ $\pm$ 0.0003 & 0.0259 $\pm$ 0.0003 \\
    Micro AUROC &  \textbf{0.6360}$^*$ $\pm$ 0.0010 & 0.5000 $\pm$ 0.0000 \\
    Macro AUROC &  \textbf{0.6527}$^*$ $\pm$ 0.0013 & 0.5000 $\pm$ 0.0000 \\
    Cross Entropy Loss &  \textbf{3.2332}$^*$ $\pm$ 0.0026 & 3.3509 $\pm$ 0.0000 \\
    
    \bottomrule
    \\
  \end{tabular}
  \caption{Phoneme classification performance. Mean and standard error of metrics over 10 seeds.}
  \footnotesize%
    $^{*}$ $p<0.01$, one‑sided exact permutation test using all 1,024 sign‑flip permutations.%
  \label{tab:phoneme_performance}
  \end{table}

%% file: 5.3_word_classification.tex
\subsection{Task 3: Word Classification}

In word classification, we segment a window of MEG data aligned to the onset of a word in the audio. Much like phoneme classification, the task is to classify the word that the neural response is associated with rather than the phoneme. 
We follow recent work in limiting classification to the set of 250 highest frequency words \citep{dascoli2024, jayalath2025cracking}.

Word classification is a task with precedent in MEG. In a recent preprint, \citet{dascoli2024} developed a method in which they encode a set of consecutive MEG windows, aligned to word onsets, with a transformer and simultaneously predict target word embeddings for these windows. The target word embeddings are extracted from the middle layer of a T5 large language model \citep{raffel2020exploring} and the word classifier is optimised for these targets using a variant of the SigLIP contrastive loss \citep{zhai2023sigmoid}. 

\citet{jayalath2025cracking} extend this work further to full sentence reconstruction using LLM-based rescoring and predictive in-filling of out-of-vocabulary words. They find that LibriBrain is the best performing dataset among the existing large speech datasets. Table \ref{tab:word-results} shows their word classification results in which they compare it to the next largest within-subject MEG dataset \citep{armeni2022}. Figure \ref{fig:scaling_all} shows that LibriBrain scales better than the \citet{armeni2022} dataset at like for like volumes of training data and continues to scale beyond it. 
This shows that word classification models trained on LibriBrain perform even better than the previous best-in-class dataset. 
We refer the reader to Appendix \ref{appendix:word-class} and especially \citet{jayalath2025cracking} for details.

%% file: tables/word_classification_performance.tex
\begin{table}[htb]
    \centering
    \begin{tabular}{lccc}
        \toprule
        \textbf{Metric} & \textbf{Random} & \citet{armeni2022} & \textbf{LibriBrain (ours)} \\
        \midrule
        Top-10 Balanced Accuracy & $0.0400 \pm 0.0000$ & $0.3261^* \pm 0.0019$ & $\mathbf{0.3621}^{*\dagger} \pm 0.0031$ \\
        \bottomrule
        \\
    \end{tabular}
    \caption{Word classification performance. Uncertainty is standard error. Results are reproduced with permission from \citet{jayalath2025cracking}. They use a vocabulary consisting of the top 250 most frequent words in each dataset. This is enough to cover 67.9\% of the story text in LibriBrain. Balanced accuracy represents the macro average over the individual accuracies of each of the words in the vocabulary. The random chance baseline is calculated as ${10}/{250} = 0.04$.}
    \label{tab:word-results}
    \footnotesize%
    $^{*}$ $p<0.01$, two‑sided permutation test using all sign‑flip permutations and $^\dagger$ two-sample $t$-test against Armeni.%
\end{table}

%% file: 6_discussion.tex
\section{Discussion}\label{sec:discussion}

The release of LibriBrain represents a significant milestone in non-invasive speech decoding research, providing, to date, the deepest within-subject MEG dataset --- 5$\times$ the next biggest \citep{armeni2022} and 25--50$\times$ most datasets. 
Unlike previous datasets that emphasise breadth across subjects, LibriBrain's depth within a single subject enables exploration of decoding limits when individual variability is controlled. This approach aligns with observations that deep, single-subject data often yields better performance than broader but shallower multi-subject datasets when total data hours are matched \citep[see e.g.~][]{dascoli2024}.
The scaling properties observed in our initial experiments align with patterns seen across machine learning domains: performance improves approximately logarithmically with increasing data volume. 
The effectiveness of averaging multiple instances of the same phoneme is particularly encouraging, as it suggests practical pathways toward useful non-invasive speech BCIs even with current technology.

Our decision to standardise data splits and establish benchmark tasks addresses a critical gap in the field. The absence of widely accepted standards has made it difficult to compare methods across studies and track progress systematically. By providing not only the data but also baseline implementations, benchmarks, as well as a Python library for easy access, we hope to lower the barrier to entry and encourage broader participation from the machine learning community.
We anticipate that LibriBrain will serve as a foundational resource, facilitating the continued expansion of non-invasive neural decoding research. Its modular design allows straightforward addition of new subjects, modalities, or tasks in future releases. Its integration with the \texttt{pnpl} library provides a unified interface for accessing multiple datasets, facilitating cross-dataset analysis and transfer learning approaches.

Our baseline experiments validate that deep neural networks achieve statistically significant decoding performance on this dataset. Additionally, we observe a clear relationship between the amount of training data and decoding accuracy, characterized by an approximately logarithmic scaling pattern similar to those seen in other machine learning domains \citep{llm_scaling_laws, image_scaling}. The particular effectiveness of averaging multiple phoneme instances suggests practical pathways toward useful non-invasive speech BCIs.

%% file: 6.0_broader_impacts.tex
\subsection{Broader Impact}

Beyond its immediate scientific contributions, LibriBrain has potential for broader impact across several domains:

\paragraph{Clinical applications.} While current non-invasive BCIs lag behind invasive approaches in performance, the scale and quality of LibriBrain could accelerate progress toward clinically viable non-invasive alternatives. These would come with the significant advantages of accessibility and risk reduction. 

\paragraph{Methodological transfer.} Techniques developed for LibriBrain may transfer to other neuroscience domains and recording modalities. Successful methods might transfer to clinical EEG \citep{jayalath2024scaling}—which is more widely available than MEG—potentially broadening impact to resource-limited settings.

\paragraph{Community building.} By establishing standardised benchmarks, public leaderboards, and providing open infrastructure, LibriBrain aims to foster a more cohesive research community around non-invasive neural decoding that may help consolidate methodological advances that are currently scattered across different research groups, accelerating overall progress in the field.

%% file: 6.1_limitations.tex
\subsection{Limitations}\label{sec:limitations}

Despite its significant scale and careful design, LibriBrain has several important limitations that should be considered when interpreting results or designing follow-up studies:

\paragraph{Single-subject design.} The deep, within-subject approach limits generalisability across subjects. It was chosen due to previous findings that decoding improves fastest when scaling data within individuals. Future extensions will incorporate multi-subject data.

\paragraph{Listening paradigm focus.} The dataset consists of MEG recordings during naturalistic listening to audiobooks, rather than imagined or overt speech. This design choice avoids muscle artefacts that would contaminate MEG signals and it enables collection of extensive high-quality data. Evidence suggests substantial overlap in neural circuits between speech perception and production \citep{Hickok2007, Pulvermuller2006}, and recent work suggests that models trained on listening data can, to some extent, transfer to covert speech tasks when decoding fMRI \citep{tang2023}. 

\paragraph{Focused language content.} The dataset includes over 50 hours of recordings predominantly from a single author (Arthur Conan Doyle) and genre (detective fiction), read by a single narrator. While this approach may limit the diversity of vocabulary, syntactic structures, and prosodic patterns compared to multi-author or multi-narrator datasets, it establishes a controlled foundation for understanding speech decoding fundamentals. This is meant to be analogous to a single-speaker ASR dataset, to minimise phonetic variance. Future releases will strategically introduce greater linguistic diversity to support broader language understanding and semantic decoding capabilities.

\paragraph{Preprocessing approach.} We applied minimal preprocessing (head motion correction, filtering, and downsampling to 250 Hz) and released data in a serialised format to maximise accessibility for machine learning researchers who may be unfamiliar with neuroimaging data, our aim being to make brain data as accessible as computer vision data in standard datasets like CIFAR-10 \citep{krizhevsky2009cifar10}. 
Some researchers may prefer access to raw recordings; to address this, we plan to release the raw BIDS-formatted data after the completion of the PNPL competition \citep{landau2025competition}. %

\paragraph{Absence of Brain-to-Text Decoding.} Some readers may question our focus on supervised prediction tasks rather than sequence-to-sequence brain-to-text (B2T) decoding. However, with the exception of \citet{tang2023} and \citet{jayalath2025cracking}, current non-invasive B2T approaches have yet to achieve word error rates below 100\%, with even the most promising results showing only marginal improvements when evaluated using character error rate or semantic similarity metrics \citep{jo2024eegtotext, yang2024neuspeech, yang2024mad, yang2024neugpt}.
To best progress the field, we believe it is essential to establish robust foundations before tackling the more complex objective of B2T. That said, LibriBrain is fully compatible with B2T translation tasks. Moreover, phoneme prediction is a critical component in successful invasive decoding pipelines \citep[e.g.][]{moses2021nejm, willett2023high}.

%% file: 7_conclusion.tex
\section{Conclusion}\label{sec:conclusion}

In this paper we introduced LibriBrain, a landmark dataset containing over 50 hours of within-subject MEG recordings during naturalistic speech comprehension. This represents the deepest non-invasive speech brain dataset to date, exceeding previous datasets by 5--50×. By providing extensive high-quality data from a single subject, LibriBrain enables the exploration of neural decoding limits when individual variability is controlled—a crucial complement to existing multi-subject datasets.

%% file: 8_ack.tex
\section{Acknowledgments}\label{sec:ack}
The authors would like to acknowledge the use of the University of Oxford Advanced Research Computing (ARC) facility in carrying out this work, \url{http://dx.doi.org/10.5281/zenodo.22558}, and the use of Hartree Centre resources. 
We also acknowledge NVIDIA for the generous contribution of additional GPUs. PNPL is supported by the MRC (MR/X00757X/1), Royal Society (RG$\backslash$R1$\backslash$241267), NSF (2314493), NFRF (NFRFT-2022-00241), and SSHRC (895-2023-1022). The doctoral research of Miran Özdogan is supported by the EPSRC grant EP/W524311/1.

%% file: 999_appendices.tex
\appendix

\section*{Appendices / Supplemental Materials}

%% file: 6.2_ethical_considerations.tex
\subsection{Ethical Considerations}\label{appendix:ethics}

The development and release of the LibriBrain dataset raise several important ethical considerations that we have carefully addressed throughout the research process:

\paragraph{Informed consent and participant privacy.} The participant provided informed consent for data collection and approved the sharing of pseudonymised data for research purposes, in accordance with the University of Oxford's ethical oversight. To protect participant privacy, we have implemented procedures to replace or remove data that directly identifies an individual participant for all released data. 

\paragraph{Open science and reproducibility.} We have deliberately chosen public-domain materials (LibriVox audiobooks, Project Gutenberg texts) and open-source tools to ensure that the entire dataset can be freely redistributed without licensing constraints. This commitment to open science extends to our analysis code, which is publicly available and documented to facilitate reproducibility. 

\paragraph{Risks of misinterpretation.} Brain decoding technologies often generate public excitement that can exceed their actual capabilities, as evidenced by recent popular press articles. We have been careful to avoiding claims that might be used to exaggerate the current state of non-invasive speech decoding. For transparency, we report multiple performance metrics and statistical comparisons to chance-level baselines. 

\paragraph{Dual-use considerations.} The primary goal of this research is to develop assistive technologies for communication in specific clinical populations for whom the technology would be a life changing. However, brain decoding techniques might have the potential to be misused for privacy invasion if applied without consent. The current state of non-invasive methods remains far from enabling such applications. Nevertheless, we emphasise that ethical guidelines for neural recording should always include informed consent and participant autonomy. 
Our aim in in releasing data and methods to help standardise the field is that there will be opportunities for researchers to come together around a strong set of ethical guardrails for any future technologies. 

\paragraph{Long-term data stewardship.} We are committed to maintaining the dataset's availability and integrity over time. The BIDS version will similarly be hosted on a standard public platform (e.g.~OSF\footnote{OSF is a free, open platform that is popular for hosting BIDS-formatted datasets (see \url{https://osf.io/}).}). 
In addition to the code being hosted on GitHub\footnote{\url{https://github.com/neural-processing-lab/frozen-pnpl}}, we have distributed the dataset through Hugging Face\footnote{\url{https://huggingface.co/datasets/pnpl/LibriBrain}}, ensuring redundancy and continuous accessibility.

%% file: 4_methods.tex
\section{Data Collection Methods}\label{sec:methods}

MEG recordings were acquired from a single, right-handed (male) volunteer with normal hearing and vision. The participant reported no history of neurological, developmental, or language-related disorders and was a native English speaker. Prior to participation, informed consent was obtained, authorising the use of anonymised data for research purposes. The study was approved by the University of Oxford Medical Sciences Interdivisional Research Ethics Committee (R90053/RE002). \\

%% file: 4.2_stimuli.tex
\subsection{Stimulus Materials}\label{sec:stimuli}

Audiobooks for the MEG experiments were sourced from LibriVox (\url{https://librivox.org/}). As in \citet{armeni2022}, we used the same recording of Sir Arthur Conan Doyle's \textit{The Adventures of Sherlock Holmes} \citep{doyle1892adventures}, plus six other audiobook recordings in the cannon of Sherlock Holmes \citep{doyle1887study, doyle1890sign, doyle1893memoirs, doyle1902hound, doyle1905return, doyle1915valley}. All were read by David Clarke, an adult male with a General British accent (see Table \ref{tab:train-audio}). 
Machine readable text for the audiobooks was recovered on Project Gutenberg (\url{https://www.gutenberg.org/}). As all audio and text used in this project are in the public domain, there are no limits on them being  openly shared for science. 

To prepare the stimuli for the MEG experiments, all audio files were converted to the uncompressed WAV format and resampled to 44.1 kHz with SoX. Our aim was to segment the audio into natural sentences or phrases, separated by pauses, and then match each segment to the appropriate text. 
This allowed us to send precise triggers to the MEG system during stimulus presentation, enabling continuous verification of the temporal alignment between MEG recordings, audio, and linguistic annotations throughout the experiment. 
Audio segmentation was achieved in phases: first automatically, using voice activity detection (VAD), then manually. 
The manual phase addressed several issues: refining segment boundaries where VAD had truncated quiet sounds (such as voiceless plosives in utterance initial position), resolving discrepancies between the written text and narrator's speech, and normalising textual representations (e.g.~converting numerical expressions like ``1066'' to match their spoken form, whether ``ten sixty-six'' or ``one thousand and sixty-six'' or something else).

Concretely, VAD was performed using custom scripts in Praat \citep{boersma2001}, identifying speech based on an intensity threshold of 59 dB and ensuring that only segments which surpassed this level and that exceeded a minimum duration of 600 ms were classified as speech. 
The resulting TextGrid annotations were then populated with corresponding text and subjected to meticulous manual correction. This labour-intensive process—requiring over 200 hours of expert human effort—significantly exceeds the typical standards for audiobook stimulus preparation. 
But it was worth it to us to reduce label noise and produce the highest-quality data possible.

%% file: 4.3_task.tex
\subsection{Experimental Design \& Procedure}\label{appendix:experimental_design}

Each session began with visually presented, self-paced instructions. After reading, the participant pressed a button to begin the experiment. A fixation cross was then displayed, followed by auditory presentation of an audiobook segment (a full chapter or partial chapter). At the end of each segment, the participant answered a comprehension question (e.g. “Where is the body of the murder victim found?”) via a two-alternative forced choice (2AFC) using a handheld response device (e.g. "A: in the bedroom", "B: in the garden"). Short breaks were allowed between the presentation of one chapter and another. \\
Instructions and fixation cues were projected on a translucent whiteboard using a DLP LED projector (ProPixx, VPixx Technologies Inc., Saint-Bruno, Canada). The stimulus computer synchronises with the MEG hardware through a parallel port to deliver triggers marking stimulus onset with high temporal precision (at a millisecond timescale). Stimulus delivery was managed via the PsychoPy toolbox \citep{peirce2007psychopy}. Auditory stimuli were delivered binaurally through tube earphones (Aero Technologies) at approximately 70 dB SPL, with minor adjustments to bass and treble based on participant preference. Responses were recorded using a MEG-compatible ResponsePixx Dual Handheld system (VPixx Technologies Inc., Saint-Bruno, Canada). \\
This protocol was repeated across multiple sessions. Each session lasted approximately three hours and typically covered from 3 to 5 audiobook chapters. Sessions were spaced at least one day apart, with no more than two months between sessions, depending on participant and experimenter availability. \\

%% file: 4.4_protocol.tex
\subsection{Data Acquisition}\label{appendix:data_aquisition}

Before data acquisition, the participant’s head shape was digitised using a Polhemus Fastrak 3D digitiser (Polhemus, Vermont, USA). This process included identifying fiducial landmarks (nasion, and left and right pre-auricular points), as well as collecting approximately 250 additional scalp, forehead, and nose surface points. Five Head Position Indicator (HPI) coils were attached to the mastoid bones and forehead to continuously monitor head position via electromagnetic induction during scanning. \\
MEG data were recorded using a MEGIN Triux™ Neo system (York Instruments Ltd., Heslington, UK), comprising 102 magnetometers and 204 orthogonal planar gradiometers. The scanner was housed within a magnetically shielded room to minimise environmental noise. Prior to entering the room, the participant was screened for metallic objects or other sources of electromagnetic interference. During scanning, the participant was seated and positioned so that their head was in close contact with the dewar. For each recording session, the participant’s head was positioned as close as possible to a standard reference location, allowing for only minimal displacement (within a few millimeters), in order to reduce variability across sessions. Instructions were given to minimise head, body, and limb movements throughout the session.
Recordings were sampled at 1000 Hz and band-pass filtered between 0.01 and 330 Hz. Eye-related artifacts were tracked using bipolar electrooculogram (EOG) electrodes—one pair placed horizontally at the outer canthi, and another vertically above and below the left eye. Cardiac activity was monitored via bipolar electrocardiogram (ECG) electrodes located on the clavicle and hip. Articulatory movements were continuously monitored with electromyography (EMG), using one electrode below the cheekbone to monitor jaw movement and another between the lower lip and chin for lip movement. Prior studies suggest that EMG signals measured from these two locations provide sufficient detail to distinguish basic phonemic contrasts \citep{gracco1994speech}. \\

%% file: 4.5_preprocessing.tex
\subsection{Minimal Preprocessing Pipeline}\label{sec:preproc}

The MEG data were minimally preprocessed to remove head movement, filter, and downsample to 250 Hz. 
First, head position information was extracted from the HPI measurements. 
Any bad channels were identified, removed, and restored using interpolation from nearby channels. %
External noise (e.g. environmental noise, stationary noise) was removed from MEG recordings offline using a Maxwell Filter software (tsss- filters; \citealt{taulu2006spatiotemporal}). We used a temporally non-extended spatial Signal Source Separation (SSS) algorithm to suppress external sources of magnetic interference.
Mains power at 50 Hz and at the 100 Hz harmonic were removed using notch filters. 
Bandpass filtering between 0.1 and 125 Hz was then applied with zero-phase two-pass Butterworth filters, to remove slow drifts and aliasing artefacts related to downsampling. 
Given the Nyquist theorem, we used 125 Hz exactly (sometimes lower values used) to keep, at least in theory, high gamma signal (despite 1/f power). 
The data were then downsampled from 1 kHz to 250 Hz, resulting in recordings of 306 channels $\times$ $T$ where $T$ was measured in 4 millisecond samples.

To ensure quality control, we visualised the events and then the PSD (power spectral density) on raw data, after notch filter, after bandpass filter, and after downsampling (see Figure \ref{fig:miniprepro} for an example). 
The choice of hyperparameters was decided based on pilot data, where we explored decoding with and without the SSS step of Maxwell Filtering and different bandpass filter range. So, the values used here reflect the best choices in pilot analyses. %

%% file: 4.6_events.tex
\subsection{Annotations/Event Files}\label{sec:events}

Annotations for each session include onset times and durations (in seconds) for event types such as silence, word (e.g.~\textit{A}, \textit{Study}, \textit{in}, \textit{Scarlet}), and phoneme (e.g.~\textit{ah}, \textit{s}, \textit{t}, \textit{ah}, \textit{d}, \textit{iy}, \textit{ih}, \textit{n}, \textit{s}, \textit{k}, \textit{aa}, \textit{r}, \textit{l}, \textit{ah}, \textit{t}) (see Table \ref{tab:example_event_file}) The annotations include other information, notably the position of the phonemes within a word. Position is indicated using the conventions from Kaldi \citep{povey2011kaldi} (\textit{B} = beginning, \textit{I} = inside,  and \textit{E} = end; the symbol \textit{S} = singleton, which is both the beginning and end of a word). 
For each MEG session (FIF file), there is a corresponding event file (TSV file). The event files can be used to specify labels for supervised tasks. The \texttt{pnpl} library comes with methods to load speech/non-speech and phonemes. As a community project, we aim to add support for other standard decoding tasks. Pull requests to the github repo are welcome.

The event files were created from the manually corrected (e.g. ~text normalised) transcripts what were used in the design of the MEG experiment. Text and audio were then force-aligned using Gentle \citep{gentle}. As the text was already aligned to short utterances, as described above (section \ref{sec:stimuli}) using VAD and over 200 hours of quality control and manual correction, the job of the forced-aligner was greatly simplified. Rather than pass entire chapters to the forced-aligner, we passed it the short utterance--text pairs. As a sanity check for the quality of the forced-aligner annotations, we examined the duration of linguistic segments of varying lengths—such as consonants, vowels, and words categorised by character count as short ($\leq 3$ characters), medium (4–6 characters), or long ($\geq 7$ characters). As expected, longer linguistic units were associated with correspondingly longer temporal annotations, providing internal validation for the temporal accuracy of the alignment (see Table \ref{tab:unit_duration}).

When segmentation using the standard configuration options failed, we explored alternative alignment strategies to improve accuracy. Specifically, enabling the "include disfluencies" option allowed the forced aligner to incorporate non-lexical vocalisations and disfluencies (e.g., “um,” “uh,” false starts) that are typically excluded from transcripts. This adjustment helped the model better capture natural variations in spontaneous speech. Additionally, enabling the "conservative" alignment option prompted the algorithm to assign word boundaries only when it had high confidence in the timing correspondence between the transcript and the audio, thereby reducing misalignments in noisy or ambiguous segments. To further address issues caused by overlapping or coarticulated speech, we ran the alignment process multiple times while systematically removing one word at a time from the transcript. This iterative method helped disentangle problematic word pairs.

Despite these precautions, the forced aligner occasionally produced sub-optimal outputs, which we identified and corrected through a labour-intensive manual inspection and adjustment process. For instance, some word occurrences were incorrectly labelled as out-of-vocabulary (OOV)—not because they were absent from the vocabulary (e.g. foreign language words, proper names, etc.), but because the forced-alignment algorithm failed to detect them for various reasons (e.g. foreign accent, excessive coarticulation, noisy pronunciation etc.). Many of these misclassified OOV instances were subsequently corrected by visually inspecting the speech spectrogram and manually annotating words and phonemes onsets and offsets using Praat \citep{boersma2001}. Following this intervention, the proportion of OOV annotations was reduced to approximately 3.5\% per book chapter. 

Finally, we expect brain responses to lag behind auditory stimuli due to neural conduction delays along the ascending auditory pathway, from the cochlea through the brainstem and thalamus to the auditory cortex and beyond \citep[e.g.~][]{schnupp2012book, parkerjones2021chapter}. However, we did not apply a fixed temporal offset (or “fudge factor”) when generating the event files. One reason is that conduction delays may vary across brain regions and processing stages. Rather than imposing a single correction, we chose to preserve the original speech timings in the event annotations. This allows users to explicitly model latencies in their analyses or to fit models that can learn temporal offsets directly from the data.

%% file: 4.7_splits.tex
\subsection{Standard Data Splits}\label{sec:data-splits}

Through many of the success stories of deep learning, it has been important to have standard data splits (e.g.~MNIST, CIFAR-10, ImageNet). This provides like for like comparisons when developing methods. 
Poor data splits can also be problematic, either to contaminate the training set or leak information. 
A potential source of leakage in neuroimaging studies is the use of data from the same set of scanning sessions for training and testing. %
Certainly, taking test data from between samples of training data can be problematic. %
Taking the last data from the day as test has been used in many cases, %
but may still leak information that would not be available in the use case scenario in which train and test took place on different days. 
Therefore, we opted for the more conservative setup, scheduling a special scan to collect all of the holdout data on a separate day from the rest of the training data in the dataset. 
This means that there are two good reasons to use the designated holdout data in future: (1) consistency in the community, and (2) strongest control against data leakage. 

The holdout data, which were acquired on a separate day from any train data, are Sherlock1 chapters 11--14. 
Specifically, we use Sherlock1 chapter 11 as the validation dataset ($\text{duration}\approx21$ minutes), %
to have the same distribution as the test data.
The standard test data are Sherlock1 chapter 12 ($\text{duration}\approx23$ minutes). 

We are holding out the data from Sherlock1 chapters 13 and 14 as evaluation sets for future machine learnign competitions ($\text{duration}\approx27 + 14$ minutes). 
This still leaves over 50 hours in this release for training alone; it leaves over 52 hours for training, validation, and test (see Table \ref{tab:splits}).

%% file: supplements/extended_dataset_stuff.tex
\section{Additional Dataset Details}\label{sec:app}

\begin{figure}[htbp]
    \centering
\includegraphics[width=\linewidth]{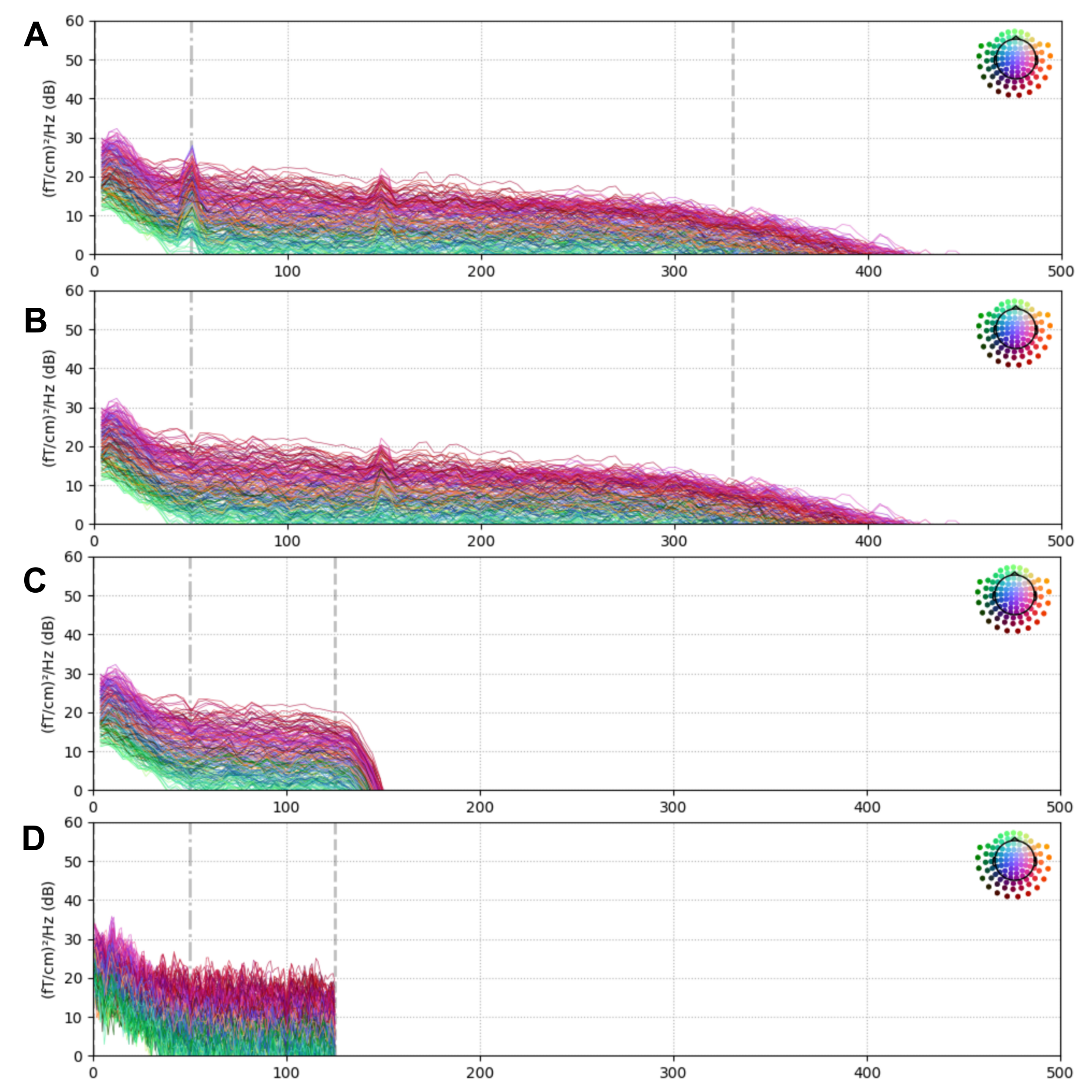}
    \caption{Minimal pre-processing example for one representative recording session. Power spectral density estimated for 204 gradiometers (colour-coded according to their spatial position, see topography in the upper right) is represented in each panel. (A) Raw data, no preprocessing. (B) Notch filter is applied, removing 50 Hz power line noise. (C) Bandpass filter 0.1-125 Hz is applied, attenuating power outside this frequency range. (D) Downsampling to 250 Hz is applied.}
    \label{fig:miniprepro}
\end{figure}

Further details on the audio books used in LibriBrain are summarised in Table~\ref{tab:train-audio}. Table~\ref{tab:example_event_file} provides an example of the event annotations provided in the dataset. Table~\ref{tab:unit_duration} presents a validation of forced alignment annotations, showing the relationship between linguistic unit length and corresponding temporal duration. Additionally, Figure~\ref{fig:miniprepro} illustrates the minimal preprocessing steps applied to a representative MEG recording session.

\input{tables/audio-datasets}

\begin{table}[htbp!]
\centering
\renewcommand{\arraystretch}{1.2} %
\rowcolors{2}{gray!10}{white} %
\begin{tabular}{|c|c|c|c|c|c|}
\hline
\rowcolor{orange!50}
\textbf{\#} & \textbf{onset} & \textbf{duration} & \textbf{type} & \textbf{segment} & \textbf{position} \\
\hline
1	& 28.772 &	1.188 &	silence & &  \\	
\hline
2	& 30.084 &	0.1 &	word &	A & \\
\hline
3	& 30.084 &	0.1 &	phoneme &	ah & S \\
\hline
4	& 30.184 &	0.37 &	word &	Study & \\
\hline
5	& 30.184 &	0.08 &	phoneme &	s & B \\
\hline
6	& 30.264 &	0.08 &	phoneme &	t & I \\
\hline
7	& 30.344 &	0.08 &	phoneme &	ah & I \\
\hline
8	& 30.424 &	0.06 &	phoneme &	d & I \\
\hline
9	& 30.484 &	0.07 &	phoneme &	iy & E \\
\hline
10	& 30.556 &	0.14 &	word	& in & \\
\hline
11	& 30.556 &	0.05 &	phoneme & ih & B \\
\hline
12	& 30.604 &	0.09 &	phoneme & n & E \\
\hline
13	& 30.696 &	0.59 &	word &	Scarlet & \\
\hline
14	& 30.696 &	0.09 &	phoneme & s & B \\
\hline
15	& 30.784 &	0.07 &	phoneme & k & I \\
\hline
16	& 30.856 &	0.11 &	phoneme & aa & I \\
\hline
17	& 30.964 & 0.01 &	phoneme & r & I \\
\hline
18	& 30.976 &	0.05 &	phoneme & l & I \\ 
\hline
19	& 31.024 &	0.07 &	phoneme & ah & I \\
\hline
20	& 31.096 &	0.19 &	phoneme & t & E \\
\hline
\end{tabular}
\caption{An example of an ‘*events.tsv’ file. Each row logs one event relative to the MEG recording session. Time values for both onset and duration are in seconds. For this example, we only show the events corresponding to one sentence extracted from the first audio book.}
\label{tab:example_event_file}
\end{table}

\begin{table}[htbp]
  \centering
  \begin{tabular}{lcc}
    \toprule
    \textbf{Linguistic unit} & \textbf{Mean} & \textbf{Standard Deviation} \\
    \midrule
    Consonants    & 0.076 s & 0.039 s \\
    Vowels      & 0.082 s & 0.042 s\\
    Short Words ($\leq 3$ characters) & 0.158 s & 0.083 s \\
    Medium Words (4--6 characters)     & 0.308 s & 0.129 s \\
    Long Words ($\geq 7$ characters)     & 0.515 s &  0.148 s \\
    \bottomrule
  \end{tabular}
  \caption{Forced-aligner validation based on annotated linguistic units duration. Longer linguistic units were associated with longer temporal annotations.}
  \label{tab:unit_duration}
\end{table}

\subsection{Phonemes colour-coding scheme}

Phonemes were grouped into perceptually and acoustically meaningful categories relevant to speech perception (see Figure~\ref{fig:teaser}). These groupings are consistent with previous literature investigating how the brain organises speech input along perceptual dimensions that are rooted in acoustic space. The rationale for choosing the colour-coding scheme applied to International Phonetic Alphabet (IPA) vowels and consonants charts is described below.

The vowels are grouped into five categories corresponding to broad acoustic similarity patterns. In particular, formant structure, diphthongisation, and vowel tenseness.

\begin{itemize}
    \item \textbf{High front vowels} $[iy, ih]$: These vowels are characterised by low F1 and high F2, distinguishing them acoustically as front and close. This group reflects fine differences in vowel height and tenseness (e.g., tense 'iy' vs. lax 'ih').
    \item \textbf{High back vowels} $[uh, uw]$: These share low F1 and lower F2 values, indicating backness and closeness. The grouping captures rounding and tongue root position differences.
    \item \textbf{Back diphthongs and mid-back vowels} $[oy, ow, ao]$: These vowels include both monophthongal ('ao') and diphthongal ('oy', 'ow') elements, but all share similar backness and rounded quality, leading to comparable acoustic transitions in F2 and F3.
    \item \textbf{Mid-front vowels} $[eh, er, ey]$: This group includes vowels with mid-height and either fronted or rhoticised qualities. The inclusion of 'er' and 'ey' reflects their transitional or complex spectral features.
    \item \textbf{Low and diphthongal vowels} $[ah, aw, aa, ae, ay]$: These vowels tend to have high F1 and variable F2 trajectories. They include open and central vowels as well as diphthongs like 'ay', capturing broad dynamic articulations.
\end{itemize}

These groupings reflect acoustic proximity and perceptual clustering \citep{miller1989vowel}, which are more informative for understanding neural responses to speech perception than articulatory descriptions alone. In the auditory cortex, vowel processing is driven by spectral characteristics \citep{mesgarani2014phonetic}, particularly formant patterns \citep{obleser2008bilateral}. This grouping schema aligns well with multidimensional scaling studies of vowel similarity and perceptual confusability, supporting their use in neural decoding and classification models of speech perception \citep{iverson1995mapping}. 

The consonants are grouped into five categories according to manner of articulation, rather than place of articulation.

\begin{itemize}
    \item \textbf{Stops} $[p, b, t, d, k, g]$: Characterised by a closure phase and a burst of release energy, stops are defined by temporal silence followed by transient noise.
    \item \textbf{Fricatives and glottal fricatives} $[f, v, th, dh, s, z, sh, zh, hh]$: Sustained high-frequency turbulence is the key acoustic cue. They form a perceptual continuum based on noise spectra and voicing.
    \item \textbf{Nasals} $[m, n, ng]$: These consonants share low-frequency energy (nasal formants) and anti-resonances.
    \item \textbf{Affricates} $[ch, jh]$: These consonants are acoustically a hybrid between stops and fricatives, with a characteristic stop-like closure followed by frication.
    \item \textbf{Approximants and liquids} $[r, w, y, l]$: These consonants have continuous airflow and formant transitions. Grouping them together reflects their similar spectral dynamics and transitional nature.
\end{itemize}

 While place of articulation is more important during speech production, it is less distinct during speech perception, especially in connected speech, where coarticulation often blurs place of articulation cues \citep{miller1955analysis}. Moreover, place of articulation cues are acoustically more variable and less reliable than manner cues \citep{stevens1978invariant}. In contrast, manner of articulation produces salient acoustic patterns that are more consistently distinguishable in the auditory signal and more likely to drive perceptual differentiation and its neural underpinnings \citep{shannon1995speech}. This is supported by psycholinguistic research on speech perception, which shows that listeners are more sensitive to differences in manner than in place, particularly in early perceptual stages \citep{phillips2000auditory}. Moreover, MEG responses to consonants often show stronger discrimination along acoustic dimensions that reflect temporal envelope (as in stops vs. fricatives) and spectral shape (as in nasals vs. fricatives), both of which map more directly onto manner than onto place of articulation \citep{liebenthal2005neural}.

%% file: tables/audio-datasets.tex
\begin{table}[htbp]
    \centering
    \begin{tabular}{|c|l|l|l|l|}
        \hline
        \textbf{Book} & \textbf{Name} & \textbf{Notes} & \textbf{Sessions} & \textbf{Hours} \\
        \hline
        1 & \href{https://librivox.org/a-study-in-scarlet-version-6-by-sir-arthur-conan-doyle/}{A Study in Scarlet} (1887) & Novel & 14* & 04:37:34 \\
        2 & \href{https://librivox.org/the-sign-of-the-four-version-3-by-sir-arthur-conan-doyle/}{The Sign of the Four} (1890) & Novel & 12 & 04:27:31 \\
        3 & \href{https://librivox.org/the-adventures-of-sherlock-holmes-version-4-by-sir-arthur-conan-doyle/}{The Adventures of Sherlock Holmes} (1892) & Short Stories & 12 & 10:56:13 \\
        4 & \href{https://librivox.org/the-memoirs-of-sherlock-holmes-by-sir-arthur-conan-doyle-2/}{The Memoirs of Sherlock Holmes} (1894) & Short Stories & 12 & 08:53:17 \\
        5 & \href{https://librivox.org/the-hound-of-the-baskervilles-version-4-by-sir-arthur-conan-doyle/}{The Hound of the Baskervilles} (1901–1902) & Novel & 15 & 06:10:32 \\
        6 & \href{https://librivox.org/the-return-of-sherlock-holmes-by-sir-arthur-conan-doyle-2/}{Return of Sherlock Holmes} (1905) & Short Stories & 14 & 11:51:17 \\ 
        7 & \href{https://librivox.org/the-valley-of-fear-version-3-by-sir-arthur-conan-doyle/}{The Valley of Fear} (1914--1915) & Novel & 14 & 06:06:17 \\ 
        \hline
        \multicolumn{4}{c}{}&\multicolumn{1}{l}{53:02:41}\\
        \multicolumn{5}{l}{\footnotesize{*Two sessions held out for competitions.}} %
    \end{tabular}
    \caption{Audiobooks available in LibriBrain}
    \label{tab:train-audio}
\end{table}

%% file: supplements/phoneme_classification.tex
\section{Additional Phoneme Classification Details}
\label{appendix:phoneme}
\subsection{Model Architecture and Hyperparameters}

\label{appendix:phoneme:hps}

\begin{figure}[htbp]
    \centering
\includegraphics[width=\linewidth]{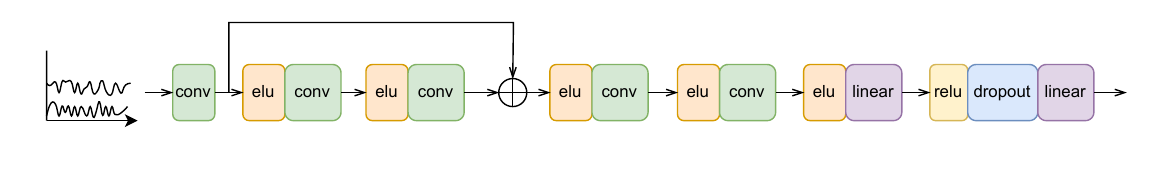}
    \caption{Convolutional Neural Network for Phoneme Classification}
    \label{fig:phoneme_architecture}
\end{figure}

We choose the SEANet architecture as a starting point because it has been successfully applied to MEG data in previous work \citep{jayalath2024scaling, jayalath2025icml}. We removed several layers because they did not increase performance. Furthermore, we add dropout \citep{dropout} which we found to increase validation performance. A visualisation of our architecture can be found in Figure~\ref{fig:phoneme_architecture}. Hyperparameters can be found in Table~\ref{tab:phoneme_hyperparameters} and \ref{tab:phoneme_architecture_hyperparameters}. We chose these based on a manual exploration. We evaluate the test partition using the model checkpoint with the best validation loss.  
For the experiments with averaged phoneme samples we change the learning rate to $0.00025$. An exception are the scaling experiments with 10 averaged input samples where we use learning rate $0.0005$.

\begin{table}[H]
  \centering
  \caption{Phoneme Classification Hyperparameters}
  \label{tab:phoneme_hyperparameters}
  \begin{tabular}{lc}
    \toprule
    \textbf{Hyperparameter} & \textbf{Value} \\
    \midrule
    Learning Rate     & 1e-4 \\
    Batch Size    & 256 \\
    Dropout Rate    & 0.5 \\
    Seeds                & [0-9] \\
    Optimizer & Adam \citep{kingma2017adammethodstochasticoptimization} \\
    Beta 1  &  0.9 \\
    Beta 2 & 0.999 \\ 
    Epsilon & 1e-08 \\
    \bottomrule
  \end{tabular}
  \end{table}
  \begin{table}[H]
  \centering
  \caption{Architecture Hyperparameters (Layer-by-Layer)}  
  \label{tab:phoneme_architecture_hyperparameters}
  \resizebox{\textwidth}{!}{%
  \begin{tabular}{llccccc}
    \toprule
    \textbf{Layer \#} & \textbf{Type}       & \textbf{In Channels} & \textbf{Out Channels} & \textbf{Kernel Size} & \textbf{Stride} & \textbf{Padding} \\
    \midrule
    1  & Conv1D          & 306 & 128 & 7  & 1 & same \\
    2a & ResNet Block – Conv1D & 128 & 128 & 3  & 1&same \\
    2b & ResNet Block – Conv1D & 128 & 128 & 1  & 1&same \\
    3  & ELU Activation  & -   & -   & -  & - & -    \\
    4  & Conv1D          & 128 & 128 & 50 & 25&none \\
    5  & ELU Activation  & -   & -   & -  & -  & -      \\
    6  & Conv1D          & 128 & 128 & 7  & 1&same \\
    7  & ELU Activation  & -   & -   & -  & -   & -     \\
    8  & Flatten         & -   & -   & -  & -   & -     \\
    9  & Linear          & 512 & 512 & -  & - & -       \\
    10 & ReLU Activation & -   & -   & -  & -   & -     \\
    11 & Dropout & - & - & - & - & - \\
    12 & Linear          & 512 & \#classes  & -  & -   & -     \\
    \bottomrule
  \end{tabular}
  }
\end{table}

\subsection{Additional Results}

\begin{figure}[htbp]
    \centering
\includegraphics[width=\linewidth]{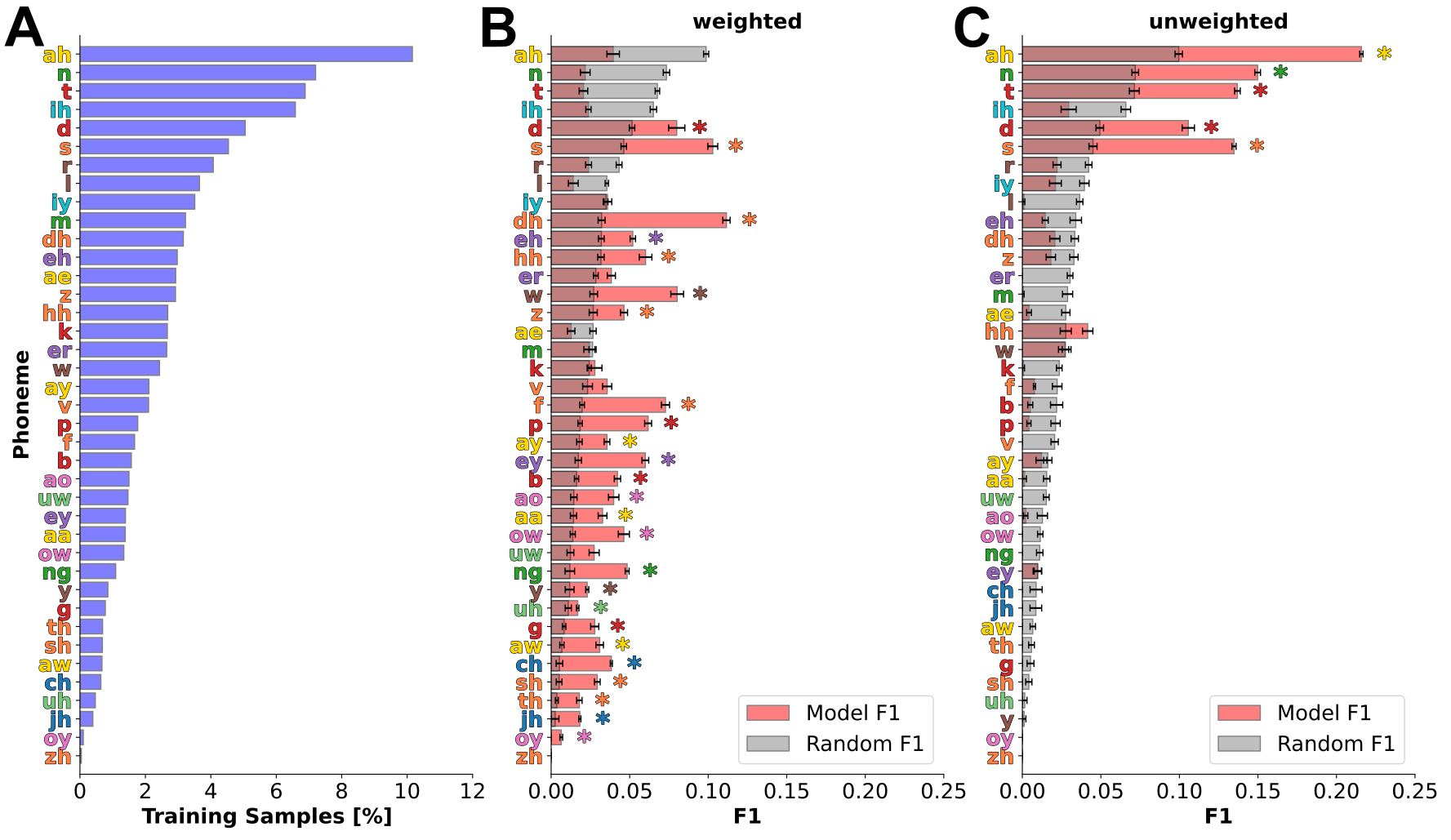}
    \caption{Per-phoneme analysis of our model's performance. (A) Distribution of training samples across phonemes. (B) Per-phoneme F1 scores when using a weighted loss function; significant differences to random baseline indicated (p < 0.01, one-sided exact permutation test). (C) Per-phoneme F1 scores as in (B), but using the default unweighted loss function.}
    \label{fig:phoneme_f1_scores}
\end{figure}

\begin{figure}[htbp]
    \centering
\includegraphics[width=0.5\linewidth]{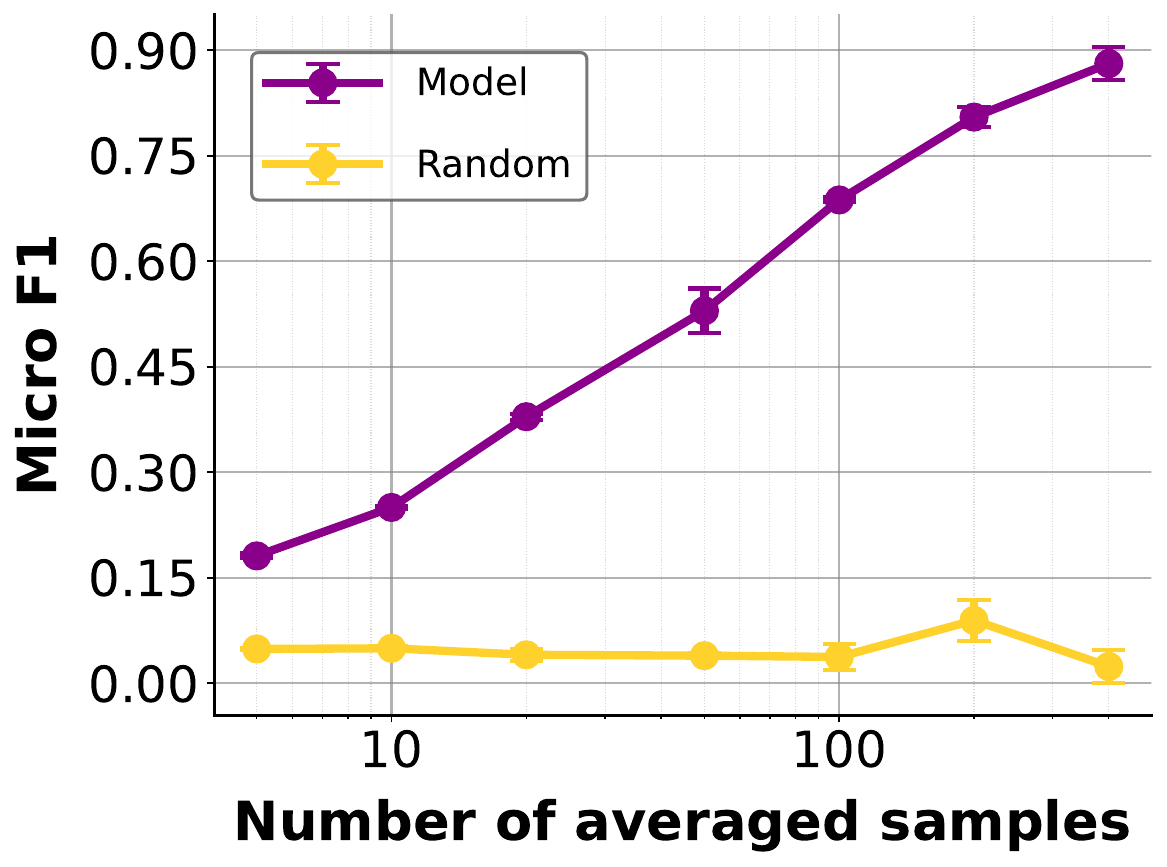}
    \caption{Effect of sample averaging on phoneme classification performance. Accuracy improves with the number of input windows averaged per phoneme.}
    \label{fig:phoneme_averaging}
\end{figure}

\begin{figure}[htbp]
    \centering
\includegraphics[width=0.85\linewidth]{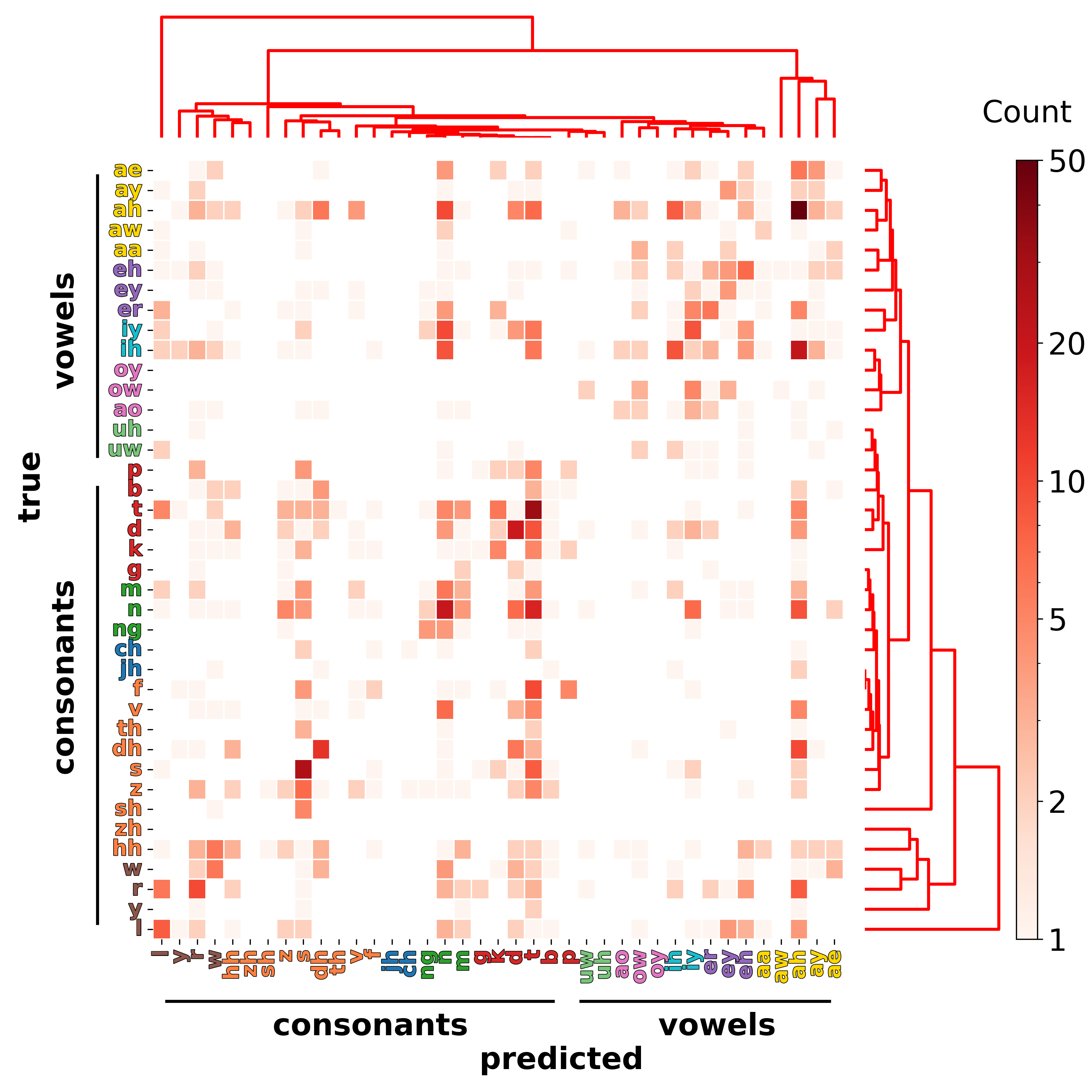}
    \caption{Confusion Matrix for the Phoneme Classification Task averaging over 10 samples for each phoneme. The distribution of predicted versus true phoneme labels is represented for 39 phonemes. Tick labels are colour-coded according to a predefined colour scheme based on the International Phonetic Alphabet (IPA) classification, grouping phonemes by place and manner of articulation: consonants and vowels. The 24 consonants are listed first, followed by 15 vowels. Dendrograms along the top and right margins represent the output of an unsupervised hierarchical clustering algorithm applied to the confusion matrix rows and columns, respectively, highlighting structural similarities in prediction patterns without affecting the phoneme order.}
    \label{fig:confusionmatrix}
\end{figure}

Here we provide additional results on phoneme classification.
In Figure~\ref{fig:phoneme_f1_scores}B-C, we show the phoneme specific F1 scores of our model. For this we use the output of the full phoneme model and for each phoneme consider the one-vs-all binary classification problem. For reference, in Figure~\ref{fig:phoneme_f1_scores}A, we show the frequencies of each phoneme in the English language. In Figure~\ref{fig:confusionmatrix}, we show the confusion matrix for the phoneme classification model predictions averaging over 10 samples for each phoneme. Furthermore, we consider the task of predicting averaged phoneme samples. For this we take the MEG window of multiple samples corresponding to the same phoneme and take their average both during model training and evaluation. In Figure~\ref{fig:phoneme_averaging} we show how performance increases with the number of samples averaged.

\subsection{Computational Requirements}
\label{appendix:phoneme_compute}

Each of our individual phoneme experiments can be replicated in under 12 hours on an NVIDIA H100 GPU. Given that we have conducted a total of 52 experiments this means that all of the phoneme results are reproducible in less than 624 GPU hours. They require no more than 64GiB of memory and less than 100GiB of storage. We conducted experiments on an internal cluster. The research project required additional compute for preliminary exploration and choosing hyperparameters. 

%% file: supplements/speech_detection.tex
\section{Additional Speech Detection Details}
\label{appendix:speech}

\subsection{Model Architecture and Hyperparameters}

\label{appendix:speech:hps}

For speech detection, we reuse the architecture that we developed for phoneme classification, as described in Appendix~\ref{appendix:phoneme:hps}. The only hyperparameter we change is the learning rate, which we set to $0.0003$ for all speech experiments based on a quick manual exploration.

\subsection{Additional Results}

During speech perception, neural responses are known to entrain to the rhythmic structure of the auditory stimulus, leading to distinct signatures in MEG recordings. One prominent effect is an increase in amplitude over temporal sensors, particularly those aligned with the location of the primary auditory cortex. This amplitude enhancement reflects stronger evoked responses to the structured acoustic input of speech and is consistent with findings that auditory cortex is highly sensitive to the temporal dynamics of speech \citep{luo2007phase}. In addition to amplitude, phase consistency across trials, as measured by inter-trial coherence (ITC), is significantly elevated during speech perception compared to non-speech intervals. This increase indicates that the phase of low-frequency neural oscillations becomes more aligned across trials, likely driven by the quasi-rhythmic syllabic structure of speech \citep{ding2014cortical, peelle2012neural}. Furthermore, power in the delta (1–4 Hz) and theta (4–8 Hz) frequency bands increases during speech perception, particularly in bilateral temporal sensors. This frequency range matches the prosodic and syllabic rhythms of natural speech and reflects the alignment of endogenous oscillatory activity with external stimulus dynamics \citep{giraud2012cortical}. \\ 

\begin{figure}[htbp]
    \centering
\includegraphics[width=\linewidth]{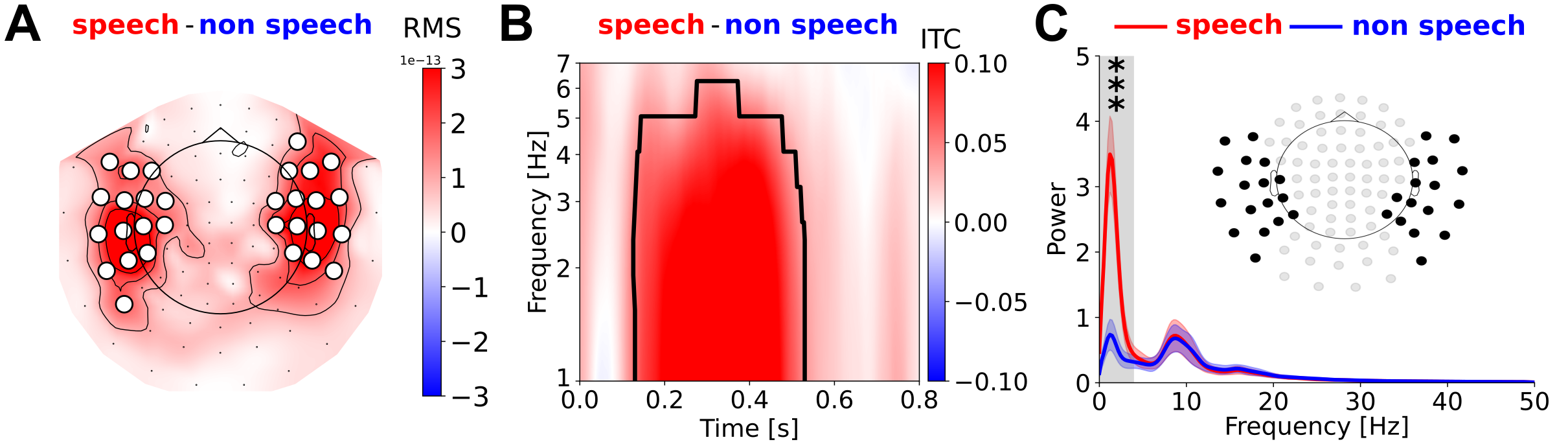}
    \caption{Differences in amplitude, phase, and power between speech and non-speech segments. (A) Topographic map of the root mean square (RMS) difference between speech and non-speech event-related potentials (ERPs), showing that speech segments exhibit higher average amplitude in temporal sensors. Statistically significant sensors are indicated by white dots. (B) Time–frequency representation of the difference in inter-trial coherence (ITC) between speech and non-speech segments. Speech segments demonstrate greater phase consistency across trials in lower frequencies (1–7 Hz; delta–theta range) in temporal sensors. Statistically significant clusters are outlined with a black contour. (C) Power spectral density (PSD) for speech and non-speech segments, revealing higher power in lower frequencies (1–7 Hz; delta–theta range) in temporal sensors during speech. Statistically significant frequency bands are highlighted with a semi-transparent grey rectangle.}
    \label{fig:speechnonspeech_analysis}
\end{figure}

To ensure the reliability and consistency of the data, we compared neural responses between speech and non-speech segments across all recording sessions—including each book and chapter. We consistently observed modulations in signal amplitude, phase, and spectral power, indicating robust differences between conditions (see Figure~\ref{fig:speechnonspeech_analysis} for an example from one representative recording session). To illustrate this, we averaged the neural responses from 1,000 randomly sampled segments, each lasting 800 milliseconds following stimulus onset. For statistical validation, we employed a bootstrapping method with replacement to estimate variability, followed by a cluster-based permutation test to identify spatio-temporal and time–frequency regions showing significant differences between speech and non-speech conditions. These neural signatures align with established findings in the literature. Importantly, while such differences are robust when using trial-averaging methods, the goal of the speech detection task is to identify these modulations at the single-trial level. This presents a substantially greater challenge, as many of the relevant neural features—particularly those related to phase and power—tend to become evident only when data are averaged across multiple trials.

\subsection{Computational Requirements}
\label{appendix:speech:compute}

Each of our individual speech detection experiments can be replicated in under 12 hours on an NVIDIA H100 GPU. Given that we have conducted a total of 31 experiments this means that all of the speech detection results are reproducible in less than 372 GPU hours. They require no more than 64GiB of memory. We conducted experiments on an internal cluster. The research project required additional compute for preliminary exploration and choosing hyperparameters. 

%% file: supplements/word_classification.tex
\section{Additional Word Classification Details}
\label{appendix:word-class}

As the word classification results are taken from \citet{jayalath2025cracking}, we restate their model description and hyperparameters here, which themselves are mostly reproduced from \citet{dascoli2024}. MEG windows are first encoded by a signal encoder. The input windows, aligned to consecutive word onsets, pass through a spatial attention \citep{defossez2023} that projects each input window channel dimension into a latent dimension of size 270 through an attention mechanism with scores derived from the $(x, y)$ positions of the sensors. Then, they are encoded through a set of dilated convolutions using the same encoder as \citet{defossez2023} which produces a temporal embedding for every time point of the input window. They take the mean over the time dimension to reduce this to a single embedding. If there are $N$ word-aligned input windows, there will be $N$ such embeddings.

Finally, these embeddings pass through a 1024-dimensional bidirectional attention transformer encoder with 16 layers, 16 heads, and rotary positional embeddings. The output embeddings of this transformer are optimised for the target middle layer embeddings of the T5 LLM (where words consisting of multiple tokens are averaged into a single target embedding).

Baseline results were obtained using the dataset released by \citet{armeni2022}, made available under the RU-DI-HD-1.0 license.

\begin{table}[H]
  \centering
  \caption{Word classification hyperparameters.}
  \label{tab:word_hyperparameters}
  \begin{tabular}{lc}
    \toprule
    \textbf{Hyperparameter} & \textbf{Value} \\
    \midrule
    Batch size / seq. length & 64 \\
    Learning rate & 1e-5 \\
    Optimiser & AdamW \citep{adamw2019} \\
    Annealing schedule & Cosine (min. 1e-6 after 50 epochs) \\
    Early stopping patience & 5 epochs \\
    Early stopping metric & Val. top-10 word class. accuracy \\
    Encoder & Brain model \citep{defossez2023} \\
    Transformer depth & 16 \\
    Transformer heads & 16 \\
    Transformer dimension & 1024 \\
    Transformer attn. dropout & 0.1 \\
    Transformer pos. emb. & Rotary \\
    \bottomrule
  \end{tabular}
  \end{table}

\subsection{Additional Results}

\begin{figure}[htbp]
    \centering
\includegraphics[width=\linewidth]{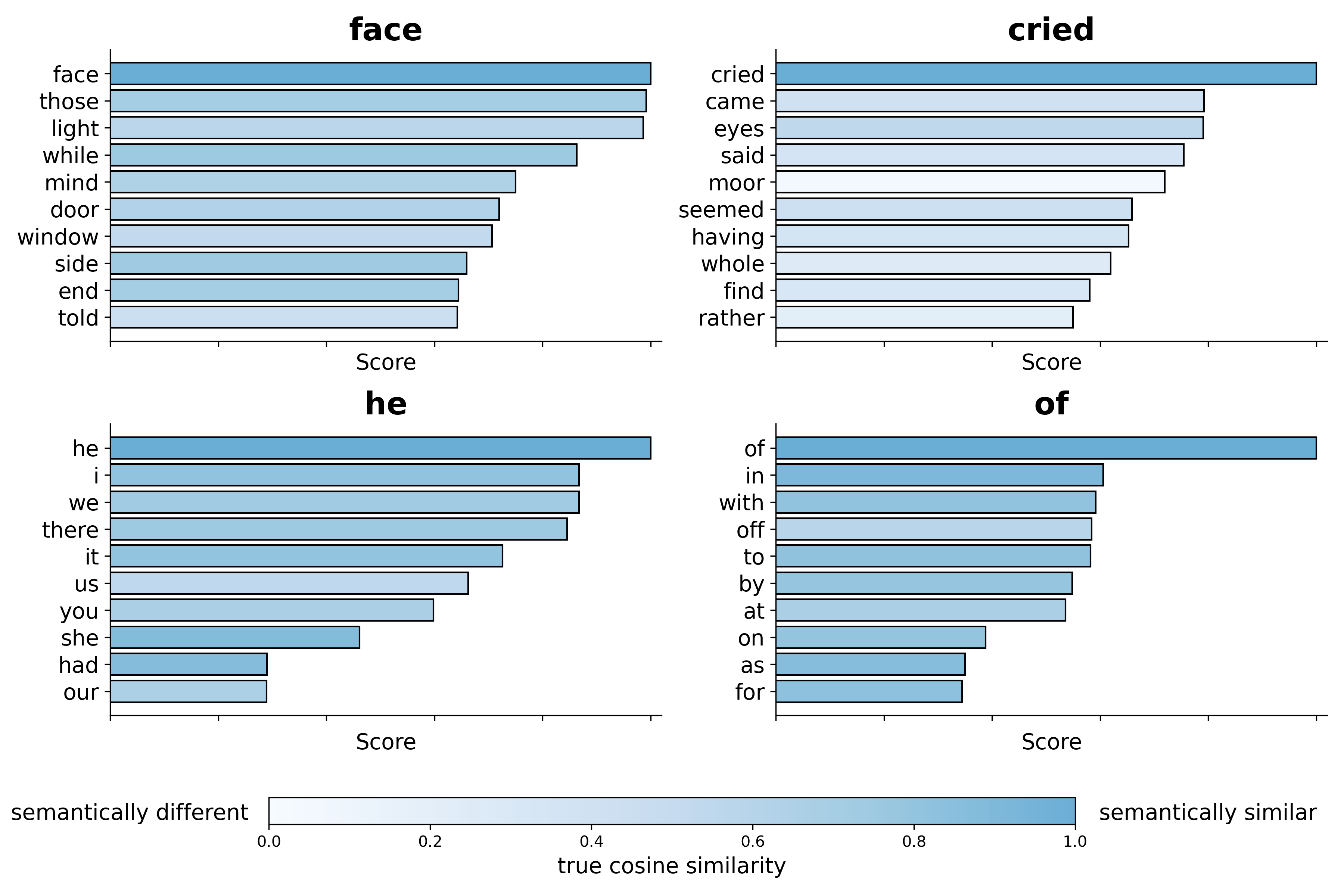}
    \caption{Top-10 model predictions for the word classification task. For illustration, four examples in which the model successfully included the correct ground truth word within its Top-10 predictions are shown. Model outputs were ranked based on softmax-normalised probabilities. To provide an interpretable view of the model's semantic performance, we visualised pairwise semantic similarity between the ground truth and the predicted words. Semantic similarity was estimated using cosine distance between 300-dimensional GloVe word embeddings.}
    \label{fig:word_semantics_prediction}
\end{figure}

\begin{figure}[htbp]
    \centering
\includegraphics[width=0.35\linewidth]{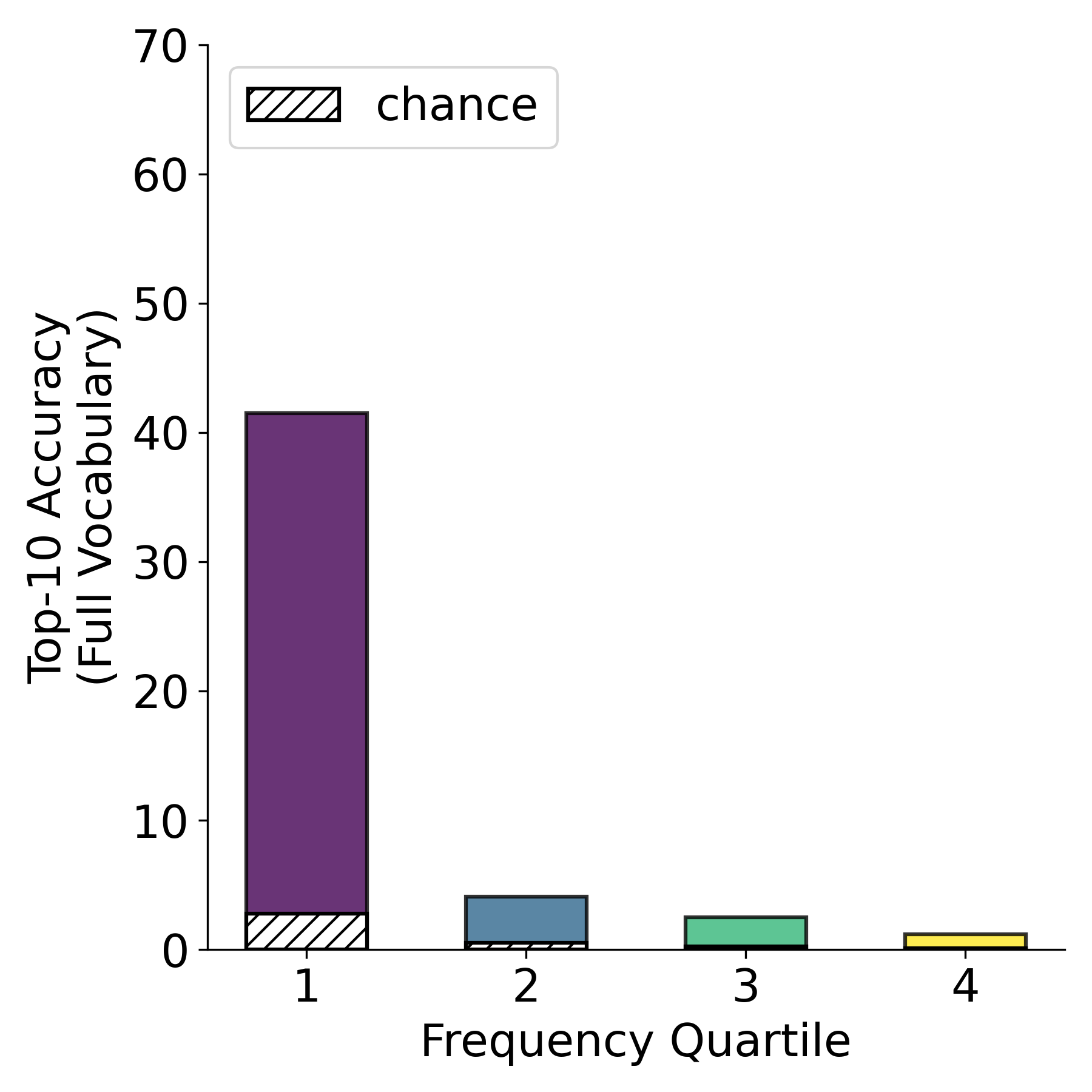}
    \caption{Impact of word frequency in the training set on classification model performance. Words in the test set were divided into four quartiles based on their frequency of occurrence in the training set, ranging from the most frequent to the least frequent. Top-10 accuracy was then computed across the full vocabulary, with each quartile reflecting the model’s performance on words belonging to that frequency range. Chance-level performance was estimated by randomly shuffling the Top-10 predictions 1,000 times to create a null distribution for comparison.}
    \label{fig:word_quartile_predictions}
\end{figure}

The word classification model output consists of predictive probabilities for each word in the vocabulary, indicating the likelihood of each word being the target word in the test set. For illustration, Figure~\ref{fig:word_semantics_prediction} represents four successful examples where the ground truth word was among the model's top ten predictions. An inspection of these top predictions across the test set revealed that classification performance is influenced by both word frequency and word type. To investigate the effect of word frequency, we divided the 250 most frequent words into four equally sized quartiles (0–25\%, 25–50\%, 50–75\%, 75–100\%) and evaluated model performance separately for each group. As shown in Figure~\ref{fig:word_quartile_predictions}B-C, accuracy drops markedly after the first quartile, indicating that higher-frequency words are more reliably predicted. To assess the influence of word type, we categorised the same set of frequent words by part-of-speech (POS): nouns, verbs, adjectives, adverbs, and function words. Performance was largely consistent across POS categories, with the exception of function words, which were predicted with notably higher accuracy. This finding aligns with the frequency effect, as function words are typically repeated more often in the training set (Figure~\ref{fig:word_quartile_predictions}A). Overall, these results underscore the importance of scaling: model performance is significantly better for words that occur frequently in the training data, highlighting the impact of the training set size on word classification accuracy.

\begin{figure}[htbp]
    \centering
\includegraphics[width=\linewidth]{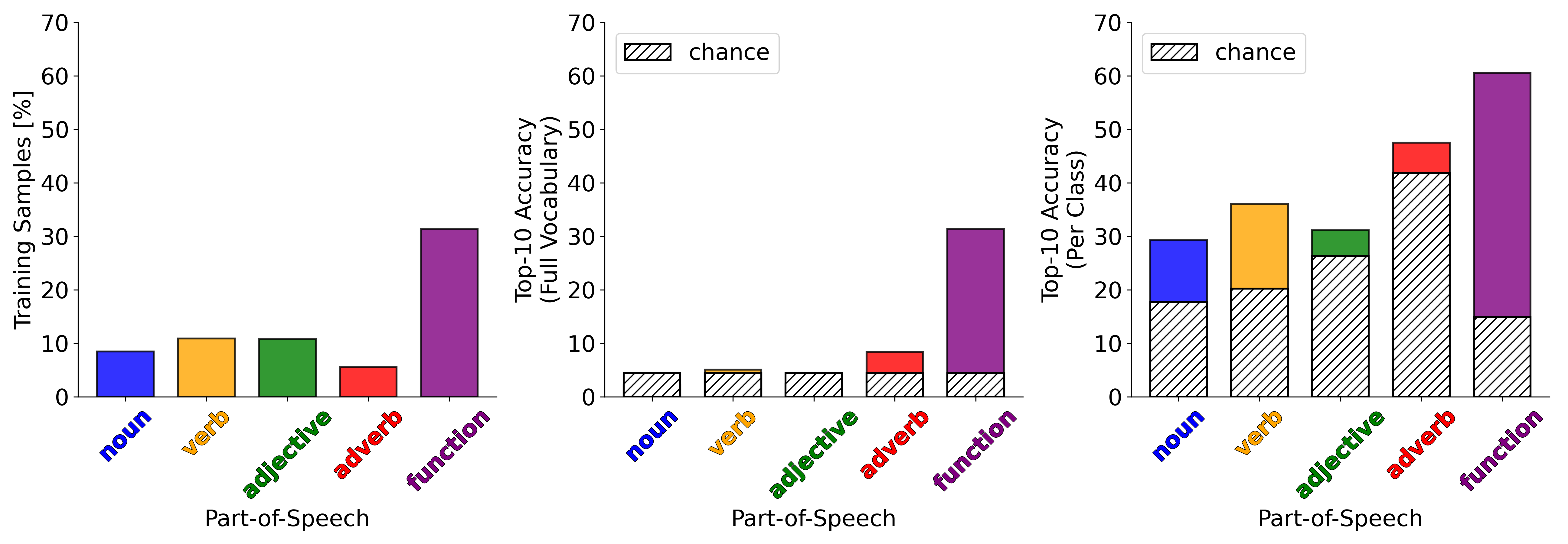}
    \caption{Effect of Part-of-Speech (POS) category on word classification model performance. (A) Distribution of training samples across POS categories, including nouns, verbs, adjectives, adverbs, and function words. (B) Top-10 accuracy computed over the full vocabulary, incorporating all instances of the 250 most frequent words in the test set. (C) Top-10 accuracy computed within each POS category, using the same 250 most frequent words grouped by their POS class. Chance-level performance was estimated by randomly permuting the model’s Top-10 predictions 1,000 times, providing a baseline for comparison.}
    \label{fig:word_pos_prediction}
\end{figure}

\begin{figure}[htbp]
    \centering
\includegraphics[width=\linewidth]{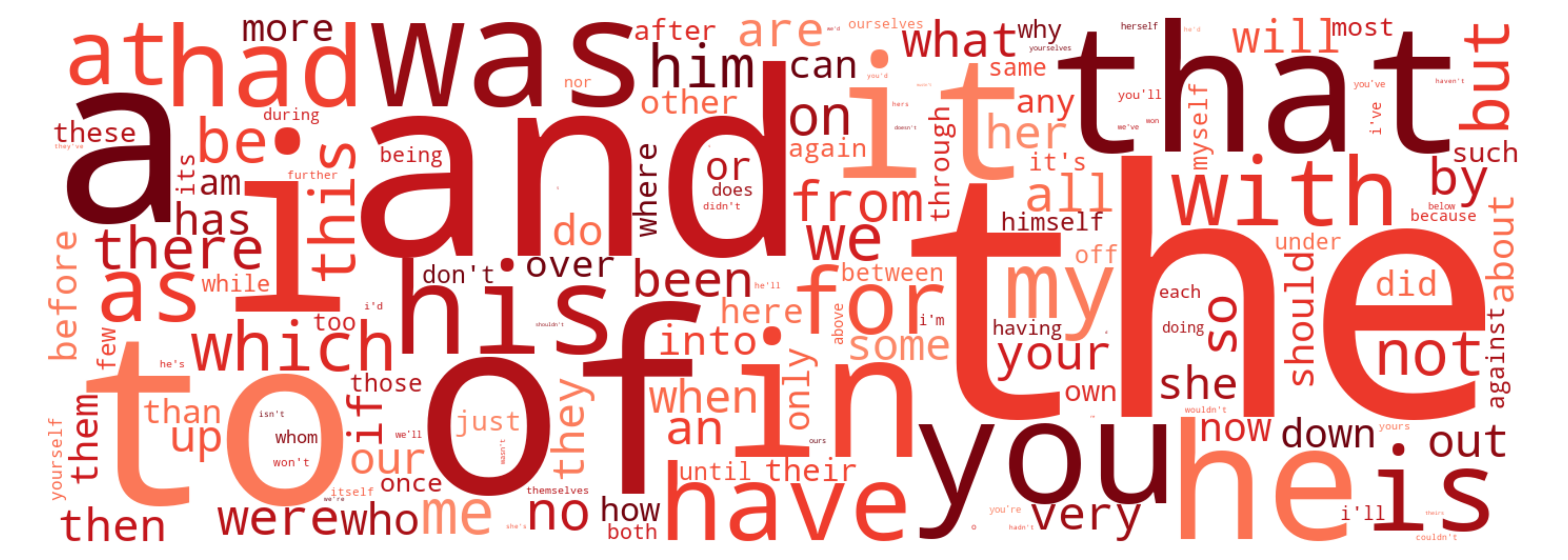}
    \caption{Top Function Words Cloud.}
    \label{fig:functionwords}
\end{figure}

\begin{table}[htbp]
\centering
\caption{Top Function Words}
\label{tab:functionwords}
\begin{tabular}{|l r | l r | l r | l r|}
\hline
Word & Freq & Word & Freq & Word & Freq & Word & Freq \\ \hline
the & 26096 & so & 1794 & about & 775 & against & 275 \\
and & 12778 & by & 1737 & before & 763 & between & 270 \\
of & 12300 & all & 1726 & only & 715 & while & 243 \\
i & 11510 & were & 1713 & am & 697 & under & 229 \\
to & 11155 & been & 1703 & here & 672 & being & 216 \\
a & 11102 & an & 1571 & did & 671 & both & 211 \\
that & 7892 & what & 1553 & any & 650 & having & 200 \\
in & 7640 & your & 1501 & how & 644 & i'll & 189 \\
it & 7326 & are & 1477 & other & 638 & whom & 173 \\
he & 7229 & very & 1390 & than & 631 & each & 173 \\
was & 6738 & her & 1387 & their & 570 & does & 169 \\
you & 6415 & if & 1371 & where & 555 & yourself & 144 \\
his & 5559 & when & 1335 & own & 490 & because & 133 \\
is & 4523 & out & 1319 & after & 471 & during & 130 \\
had & 4213 & then & 1254 & such & 470 & nor & 123 \\
have & 3822 & she & 1227 & these & 465 & i've & 108 \\
with & 3677 & will & 1211 & through & 431 & above & 108 \\
my & 3659 & up & 1204 & just & 428 & doing & 101 \\
as & 3327 & who & 1167 & once & 393 & i'm & 100 \\
for & 3278 & they & 1148 & most & 393 & ourselves & 92 \\
at & 3271 & our & 1142 & again & 386 & you'll & 80 \\
which & 3132 & some & 1137 & why & 368 & further & 73 \\
we & 2828 & has & 1131 & himself & 365 & won't & 71 \\
but & 2789 & do & 1093 & off & 345 & itself & 71 \\
me & 2538 & into & 1044 & until & 339 & yours & 62 \\
not & 2536 & or & 959 & too & 331 & he's & 61 \\
this & 2508 & now & 946 & its & 330 & didn't & 60 \\
be & 2465 & down & 923 & don't & 316 & themselves & 60 \\
him & 2303 & them & 877 & myself & 306 & i'd & 55 \\
there & 2235 & can & 874 & few & 299 & you've & 43 \\
from & 2080 & more & 863 & those & 299 & we'll & 42 \\
on & 1968 & should & 855 & same & 289 & herself & 39 \\
no & 1891 & over & 818 & it's & 285 & you're & 38 \\
\hline
\end{tabular}
\end{table}

\begin{figure}[htbp]
    \centering
\includegraphics[width=\linewidth]{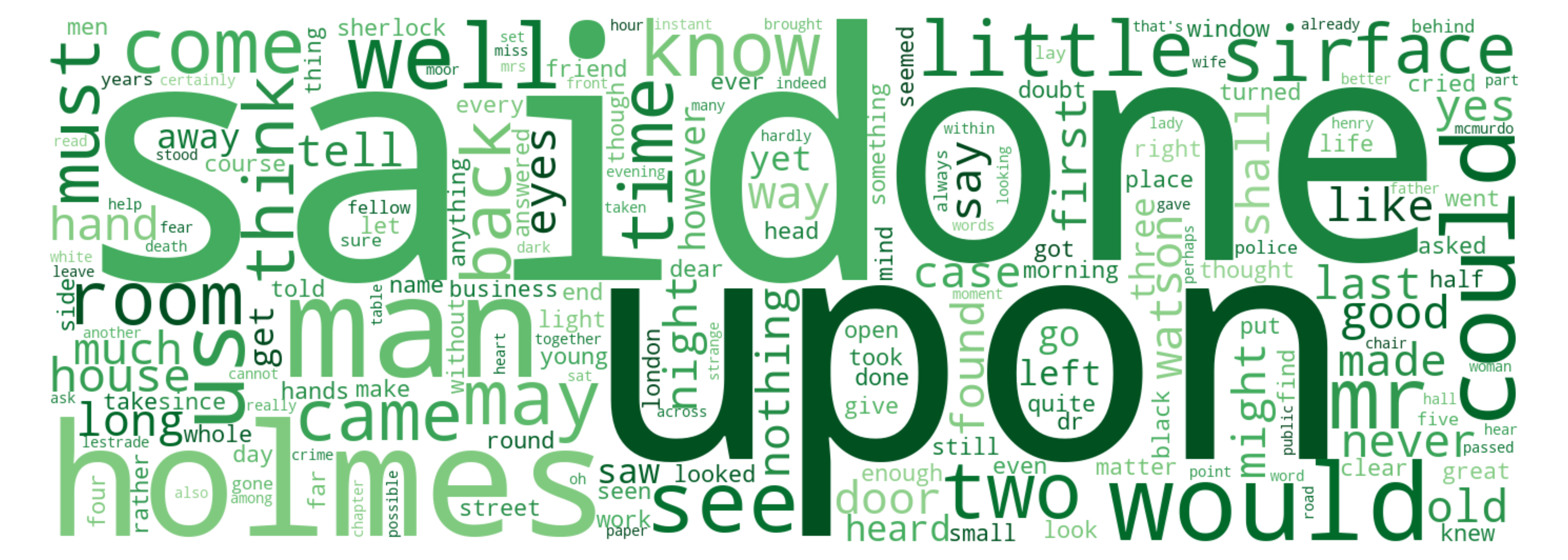}
    \caption{Top Content Words Cloud.}
    \label{fig:contentwords}
\end{figure}

\begin{table}[htbp]
\centering
\caption{Top Content Words.}
\label{tab:contentwords}
\begin{tabular}{|l r | l r | l r | l r|}
\hline
Word & Freq & Word & Freq & Word & Freq & Word & Freq \\ \hline
said & 2226 & however & 427 & looked & 276 & also & 203 \\
upon & 1893 & go & 420 & half & 274 & evening & 202 \\
one & 1860 & saw & 402 & life & 273 & help & 202 \\
holmes & 1846 & left & 399 & mind & 270 & already & 201 \\
man & 1533 & yet & 396 & knew & 269 & death & 201 \\
would & 1501 & away & 396 & turned & 265 & sat & 200 \\
could & 1320 & three & 389 & course & 262 & across & 198 \\
us & 1088 & get & 385 & seemed & 260 & hardly & 198 \\
well & 1041 & thought & 383 & even & 258 & table & 198 \\
mr & 975 & matter & 378 & though & 252 & paper & 198 \\
see & 947 & cried & 374 & anything & 248 & moor & 197 \\
two & 816 & find & 373 & london & 246 & miss & 196 \\
little & 768 & make & 371 & doubt & 245 & oh & 195 \\
sir & 765 & asked & 369 & men & 245 & mcmurdo & 195 \\
may & 762 & round & 364 & went & 244 & passed & 193 \\
come & 754 & great & 362 & answered & 243 & mrs & 193 \\
know & 732 & morning & 357 & dear & 243 & lady & 193 \\
time & 697 & day & 356 & work & 242 & moment & 192 \\
room & 693 & end & 352 & open & 239 & gave & 190 \\
came & 637 & take & 351 & told & 231 & read & 189 \\
must & 635 & young & 351 & rather & 228 & looking & 188 \\
think & 614 & every & 349 & four & 228 & hall & 188 \\
face & 598 & right & 348 & clear & 228 & leave & 187 \\
back & 597 & still & 345 & whole & 226 & indeed & 186 \\
house & 583 & friend & 344 & black & 226 & taken & 186 \\
watson & 560 & sherlock & 337 & since & 225 & white & 186 \\
way & 555 & took & 331 & dr & 225 & father & 185 \\
good & 544 & done & 327 & business & 224 & another & 184 \\
hand & 532 & side & 323 & thing & 222 & better & 184 \\
never & 526 & head & 319 & police & 221 & road & 184 \\
last & 512 & small & 312 & years & 220 & hear & 181 \\
night & 510 & give & 310 & lay & 220 & chair & 181 \\
door & 500 & look & 310 & fellow & 218 & cannot & 180 \\
case & 490 & quite & 310 & always & 217 & set & 178 \\
might & 490 & let & 307 & behind & 217 & lestrade & 178 \\
nothing & 489 & street & 307 & gone & 216 & words & 178 \\
much & 485 & light & 307 & sure & 214 & fear & 178 \\
long & 482 & hands & 304 & five & 214 & ask & 176 \\
shall & 481 & ever & 304 & certainly & 213 & hour & 175 \\
say & 471 & put & 302 & many & 209 & part & 173 \\
made & 467 & name & 302 & together & 208 & strange & 173 \\
like & 460 & place & 300 & word & 206 & among & 173 \\
first & 456 & seen & 298 & woman & 206 & really & 172 \\
found & 454 & window & 297 & instant & 205 & front & 172 \\
old & 447 & enough & 293 & point & 205 & perhaps & 170 \\
eyes & 445 & something & 293 & brought & 204 & heart & 167 \\
yes & 441 & without & 287 & within & 204 & public & 166 \\
tell & 438 & far & 286 & dark & 204 & wife & 166 \\
heard & 429 & got & 284 & stood & 204 & crime & 165 \\
\hline
\end{tabular}
\end{table}

\subsection{Computational Requirements}
\label{appendix:word_compute}

Each of our individual word classification experiments can be replicated in under 12 hours on an NVIDIA V100 GPU. Given that we have conducted a total of 30 experiments this means that all of the word classification results are reproducible in less than 360 GPU hours. They require no more than 64GiB of memory. We conducted experiments on an internal cluster. The research project required additional compute for preliminary exploration and choosing hyperparameters.